\pdfoutput=1

\documentclass[11pt]{article}

\usepackage{acl}

\usepackage{times}
\usepackage{latexsym}

\usepackage[T1]{fontenc}

\usepackage[utf8]{inputenc}

\usepackage{microtype}

\usepackage{inconsolata}

\usepackage{graphicx}
\usepackage{amsmath}
\usepackage{amssymb}
\usepackage{color}
\usepackage{booktabs}
\usepackage{multirow}
\usepackage{enumitem}
\usepackage{arydshln}

\newcommand{\TABLEFONT}{\small}
\newcommand{\cdashlinelr}[1]{
  \noalign{\vskip 2pt}
  \cdashline{#1}[2pt/2pt]
  \noalign{\vskip 4pt}}

\title{Lost in the Source Language: How Large Language Models Evaluate the Quality of Machine Translation}

\author{
    Xu Huang$^{1}\thanks{\hspace{0.2em} Work was done during internship at Tencent AI Lab.}$ \quad
    Zhirui Zhang$^{2\dag}$ \quad Xiang Geng$^{1}$ \\ 
    {\bf Yichao Du}$^{3}$ \quad {\bf Jiajun Chen}$^{1}$ \quad {\bf Shujian Huang}$^{1}$\thanks{\hspace{0.2em} Corresponding authors} \\
    $^{1}$National Key Laboratory for Novel Software Technology, Nanjing University \\ $^{2}$Tencent AI Lab \quad  $^{3}$University of Science and Technology of China \\
    $^{1}$\texttt{\{xuhuang,gx\}@smail.nju.edu.cn}, \texttt{\{chenjj,huangsj\}@nju.edu.cn} \\
    $^{2}$ \texttt{zrustc11@gmail.com} \quad $^{3}$ \texttt{duyichao@mail.ustc.edu.cn}
}

\begin{document}
\maketitle
\begin{abstract}
This study investigates how Large Language Models (LLMs) leverage source and reference data in machine translation evaluation task, aiming to better understand the mechanisms behind their remarkable performance in this task.
We design the controlled experiments across various input modes and model types, and employ both coarse-grained and fine-grained prompts to discern the utility of source versus reference information.
We find that reference information significantly enhances the evaluation accuracy, while surprisingly, source information sometimes is counterproductive, indicating LLMs' inability to fully leverage the cross-lingual capability when evaluating translations.
Further analysis of the fine-grained evaluation and fine-tuning experiments show similar results.
These findings also suggest a potential research direction for LLMs that fully exploits the cross-lingual capability of LLMs to achieve better performance in machine translation evaluation tasks.

\end{abstract}

\section{Introduction}

The last decade has witnessed significant development in Neural Machine Translation (NMT) \cite{DBLP:journals/corr/BahdanauCB14,DBLP:conf/nips/VaswaniSPUJGKP17,hassan2018achieving}.
As the quality of machine translations has been improved, it becomes more challenging and critical for automatic translation evaluation.
The recent study~\cite{DBLP:conf/wmt/FreitagRMLSAKFLM22} calls for stopping using BLEU~\cite{DBLP:conf/acl/PapineniRWZ02}, a traditional metric, as it is not reliable for high-quality translations and has a lower correlation with human judgements.
Neural metrics based on pre-trained language models~\cite{DBLP:conf/naacl/DevlinCLT19,DBLP:journals/corr/LiuOGDJ19,DBLP:conf/acl/ConneauKGCWGGOZ20}, such as COMET~\cite{DBLP:conf/emnlp/ReiSFL20}, show a higher correlation with human judgments.
However, these neural metrics only provide a score lacking interpretability and still exhibit robustness issues hard to detect~\cite{DBLP:conf/acl/YanWZHCW23}.
Recently, Large Language Models (LLMs)~\cite{DBLP:journals/corr/gpt4,Touvron2023Llama2O,DBLP:conf/emnlp/WangLJZY0T23} have also been used as translation evaluators.
GEMBA~\cite{DBLP:conf/eamt/KocmiF23} presents that GPT-4 can achieve state-of-the-art performance in system-level assessment.
While LLMs show remarkable performance in translation evaluation tasks, the reasons underlying their success have not been thoroughly investigated.

\begin{table}[tb]
    \TABLEFONT
    \centering
    \setlength{\tabcolsep}{4pt}
    \begin{tabular}{ccccc}
        \toprule
        \multirow{2}{*}{Input Mode} & Accuracy & \multicolumn{3}{c}{Kendall's $\tau$} \\
        \cmidrule(lr){2-2}\cmidrule(lr){3-5}
         & All LPs & En-De & Zh-En & En-Ru \\
        \midrule
        T & 0.759 & 0.181 & 0.228 & 0.195 \\
        S-T & 0.876 & 0.212 & 0.220 & 0.219 \\
        R-T & \textbf{0.891} & \textbf{0.284} & \textbf{0.286} & \textbf{0.253} \\
        S-R-T & 0.876 & 0.255 & 0.274 & 0.211 \\
        \bottomrule
    \end{tabular}
    \vspace{-5pt}
    \caption{The system-level accuracy and segment-level Kendall's $\tau$ correlation of ChatGPT when using different inputs.}
    \label{tab:init_exp}
\end{table}

In this paper, we take the further step to investigate how LLMs leverage source and reference information in evaluating translation in both coarse-grained and fine-grained settings, bringing better understanding of the working mechanism of LLMs.
Four input modes are defined, each of which exposes different information to LLMs.
These include Translation-only (T), Source-Translation (S-T), Reference-Translation (R-T) and Source-Reference-Translation (S-R-T) modes.
We first instruct both open and closed LLMs to predict coarse-grained quality scores using GEMBA prompt, but given different information, namely sources and references.
While references significantly improve the system-level accuracy and segment-level correlations, we surprisingly find that the source information is sometimes counterproductive.
For example, as shown in Table~\ref{tab:init_exp}, ChatGPT\footnote{We use \textit{gpt-3.5-turbo-0613} in our experiments.} achieves the best performance in the R-T mode, while S-R-T mode leads to performance degradation.
This indicates LLMs' inability to fully leverage cross-lingual capabilities when evaluating translation sentences, despite ChatGPT's impressive performance in multi-lingual or translation tasks.

In addition to the superficial score prediction, we further explore the fine-grained error detection to better understand the cross-lingual ability.
Our fine-grained experiments confirm the above observations, where we follow AutoMQM~\cite{DBLP:journals/corr/FernandesDFRMNGCFF23} method to predict error spans and categories for translation sentences and study the performance of LLMs with different input modes.
We also conduct a comprehensive meta-evaluation for the error span and error category, alongside executing a critical error detection task.
The findings suggest that LLMs struggle to fully utilize the source information for translation evaluation.

Lastly, we examine the effect of fine-tuning, which makes deeper modifications to the model, with Multidimensional Quality Metrics (MQM) data~\cite{DBLP:journals/tacl/FreitagFGRTM21}.
Although fine-tuning can greatly improve the model's performance of translation evaluation, the negative impact of source sentence still exists.
These experimental results reveal the limitation of current LLMs on machine translation evaluation tasks and suggest a potential research direction that fully exploits the cross-lingual capability of LLMs to achieve better performance.

Overall, our main contributions are as follows:
\begin{itemize}
\setlength{\itemsep}{0pt}
    \item To the best of our knowledge, we are the first to explore the working mechanism of LLMs in evaluating translation by testing the importance of sources and reference information.
    \item We conduct extensive experiments to discern the utility of source versus reference information through various aspects, suggesting that LLMs are unable to fully utilize the cross-lingual capability to evaluate translation sentences.
    \item We provide an in-depth analysis of translation error detection. Our code and data would be released for the research community to promote the development of LLMs in automatic translation estimation tasks.\footnote{Code and data are available at \url{https://github.com/NJUNLP/lost_in_the_src}.}
\end{itemize}

\begin{figure*}[h]
    \centering
    \includegraphics[width=.95\linewidth]{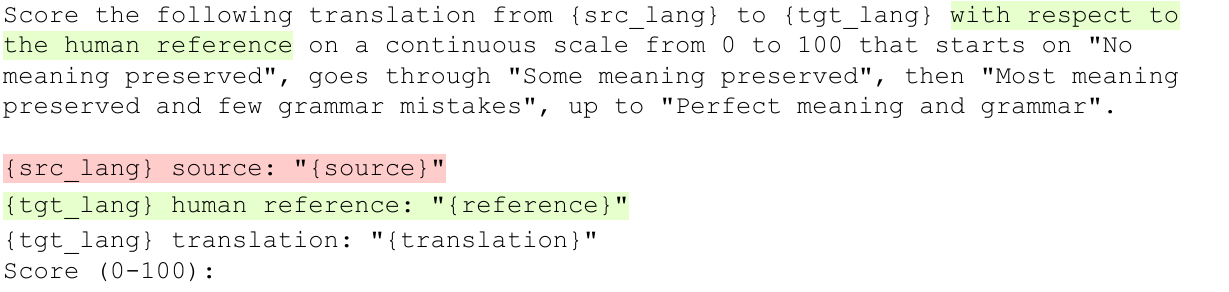}
    \vspace{-5pt}
    \caption{The GEMBA-SQM prompt template. The green parts will be included in the prompt if the reference information is given. Similarly, the red part will be in the prompt if the source is given. Detailed prompts can be found in Appendix~\ref{sec:detailofgemba}.}
    \label{fig:gemba-sqm}
\end{figure*}

\section{Related Work}
\paragraph{Automatic Translation Evaluation.}
Automatic evaluation has been a crucial and tough problem along with the development of machine translation.
Traditional metrics like BLEU~\cite{DBLP:conf/acl/PapineniRWZ02}, METEOR~\cite{DBLP:conf/acl/BanerjeeL05} and chrF~\cite{DBLP:conf/wmt/Popovic15} rely heavily on n-gram matching algorithms.
Despite their past success, they fail to keep pace with the increasing performance of translation models that generate semantically correct translations.

Neural metrics leverage the meaningful representations of pre-trained language models to evaluate translations.
BertScore~\cite{DBLP:conf/iclr/ZhangKWWA20} and MoverScore~\cite{DBLP:conf/emnlp/ZhaoPLGME19} calculate similarity using these representations, while generation-based methods like Prism~\cite{DBLP:conf/emnlp/ThompsonP20} and BartScore~\cite{DBLP:conf/nips/YuanNL21} assess text quality through generation tasks, conditioned on sources or references.
Learned metrics such as COMET~\cite{DBLP:conf/emnlp/ReiSFL20}, BLEURT~\cite{DBLP:conf/acl/SellamDP20}, UniTE~\cite{DBLP:conf/acl/0004LYZCWC22} and \textsc{xCOMET}~\cite{Guerreiro2023xCOMETTM} apply neural networks to predict quality scores in a supervised manner.
Particularly, UniTE suggests that the interaction between source and hypothesis may have an adverse effect.
Although UniTE performs best with the S-R-T mode, our work shows this is not always the case for LLMs.

LLMs that can follow the instructions of evaluation tasks are also used to evaluate translations.
GEMBA~\cite{DBLP:conf/eamt/KocmiF23} shows that GPT-4, when asked to directly generate a quality score, is the state-of-the-art translation evaluator at that time in the system-level assessment.
EAPrompt~\cite{DBLP:journals/corr/LuQDXT23}, AutoMQM~\cite{DBLP:journals/corr/FernandesDFRMNGCFF23}, and GEMBA-MQM~\cite{DBLP:journals/corr/KocmiF23} endeavor to instruct LLMs in detecting translation errors meticulously and labelling the error category and severity.
Despite their remarkable performance, the reasons underlying their success have not been thoroughly investigated. 

\begin{table*}[t]
    \TABLEFONT
    \centering
    \begin{tabular}{lcccccccc}
    \toprule
         \multirow{2}{*}{Model}& \multirow{2}{*}{Mode}  & All LPs & \multicolumn{2}{c}{En-De} & \multicolumn{2}{c}{Zh-En} & \multicolumn{2}{c}{En-Ru} \\
         \cmidrule{3-9}
         &  & Acc. & $\tau$ & $\rho$ & $\tau$ & $\rho$ & $\tau$ & $\rho$ \\
         \midrule
         \multirow{4}{*}{GPT-3.5-turbo}& T & 0.759 & 0.181 & 0.153 & 0.228 & 0.157 & 0.195 & 0.169 \\
         & S-T & 0.876 & 0.212 & 0.242 & 0.220 & 0.219 & 0.219 & 0.186 \\
         & R-T & \textbf{0.891} & \textbf{0.284*} & 0.280 & \textbf{0.286*} & 0.230 & \textbf{0.253*} & \textbf{0.217*} \\
         & S-R-T & 0.876 & 0.255 & \textbf{0.285} & 0.274 & \textbf{0.248*} & 0.211 & 0.196 \\
         \cdashlinelr{1-9}
         \multirow{4}{*}{Llama2-7B-Chat} & T & 0.620 & 0.052 & 0.036 & 0.156 & 0.195 & 0.042 & 0.054 \\
         & S-T & 0.599 & \underline{-0.010} & \underline{-0.037} & \underline{0.093} & \underline{0.121} & \underline{0.008} & \underline{0.003} \\
         & R-T & \textbf{0.788} & \textbf{0.217*} & \textbf{0.200*} & 0.284 & 0.260 & 0.213 & 0.177 \\
         & S-R-T & 0.748 & 0.187 & 0.173 & \textbf{0.290} & \textbf{0.277*} & \textbf{0.222} & \textbf{0.196*} \\
         \cdashlinelr{1-9}
         \multirow{4}{*}{Llama2-13B-Chat} & T & 0.675 & 0.000 & 0.003 & 0.034 & 0.03 & 0.032 & 0.029 \\
         & S-T & 0.591 & 0.041 & 0.028 & 0.056 & 0.041 & 0.084 & 0.038 \\
         & R-T & \textbf{0.701} & 0.107 & 0.100 & \textbf{0.104*} & \textbf{0.097*} & \textbf{0.108} & \textbf{0.105} \\
         & S-R-T & 0.650 & \textbf{0.108} & \textbf{0.109} & 0.053 & 0.055 & \textbf{0.108} & 0.102 \\
         \cdashlinelr{1-9}
         \multirow{4}{*}{Llama2-70B-Chat} & T & 0.737 & 0.148 & 0.105 & 0.215 & 0.177 & 0.220 & 0.145 \\
         & S-T & 0.807 & \underline{0.126} & 0.123 & \underline{0.194} & \underline{0.153} & \underline{0.134} & \underline{0.126} \\
         & R-T & \textbf{0.887*} & \textbf{0.241*} & \textbf{0.221*} & \textbf{0.271*} & \textbf{0.228*} & \textbf{0.222} & \textbf{0.160*} \\
         & S-R-T & 0.843 & 0.167 & 0.180 & 0.250 & 0.197 & 0.178 & 0.103 \\
         \cdashlinelr{1-9}
         \multirow{4}{*}{Mistral-7B-Instruct} & T & 0.726 & 0.108 & 0.079 & 0.232 & 0.211 & \textbf{0.228*} & \textbf{0.160} \\
         & S-T & 0.646 & \underline{0.063} & \underline{0.052} & \textbf{0.238} & \underline{0.190} & \underline{0.180} & \underline{0.131} \\
         & R-T & \textbf{0.796} & 0.123 & 0.119 & 0.228 & 0.213 & 0.158 & 0.118 \\
         & S-R-T & 0.770 & \textbf{0.157*} & \textbf{0.143*} & 0.237 & \textbf{0.228*} & 0.170 & 0.146 \\
         \midrule
         COMET-22 & / & 0.839 & 0.368 & 0.512 & 0.428 & 0.585 & 0.400 & 0.469 \\
         BLEU & / & 0.708 & 0.169 & 0.193 & 0.145 & 0.175 & 0.140 & 0.160 \\
         chrF & / & 0.734 & 0.214 & 0.231 & 0.147 & 0.154 & 0.168 & 0.168 \\
         \bottomrule
    \end{tabular}
    \vspace{-5pt}
    \caption{The system-level accuracy and segment-level Kendall's $\tau$ and Pearson $\rho$ correlations of different models with different input modes on WMT22 test set. Bold scores indicate the highest values, while asterisks mark the significantly highest among the four input modes. The underlined S-T mode scores are significantly lower than the T mode scores.}
    \label{tab:coarse}
\end{table*}

\paragraph{LLM-based Text Evaluation.}
Evaluating text quality is a challenging problem even for humans.
LLMs with broad world knowledge and expertise in linguistics, like ChatGPT, have been used as natural language evaluators.
GPTScore~\cite{Fu2023GPTScoreEA} utilizes the conditional probability for text evaluation.
Other works~\cite{Wang2023IsCA,Liu2023GEvalNE,Wang2023LargeLM,Chan2023ChatEvalTB,Liu2023CalibratingLE} prompt ChatGPT to generate evaluations for natural texts using various techniques.
InstructScore~\cite{Xu2023INSTRUCTSCORETE} proposes an explainable metric by fine-tuning the Llama-7B with data generated from GPT-4 and self-generated outputs, proving the feasibility of using open, smaller models for evaluation purposes.
This study primarily focuses on employing LLMs for assessing translations, with a major emphasis on the cross-lingual ability.
However, the insights gained from our research may extend to other NLG evaluation tasks.
LLMs demonstrate a greater ability to capitalize on reference information, while they may face challenges in effectively utilizing task input like source information.

\section{Coarse-grained Score Prediction}
We first investigate how LLMs leverage the source and reference information in conducting coarse-grained evaluations of translation quality via score prediction.
We adopt the GEMBA-SQM prompt template~\cite{DBLP:conf/eamt/KocmiF23}, as shown in Figure~\ref{fig:gemba-sqm}.
The inclusion of source and reference information in the prompt varies based on the selected input mode.
For instance, in the R-T mode, the source text in red is omitted from the prompt.
We assess the efficacy of LLMs across four distinct input modes, examining their impact on the model's performance.

\subsection{Experimental Setup}
\paragraph{Data.} We use the test set from WMT22 Metric Shared Task~\cite{DBLP:conf/wmt/FreitagRMLSAKFLM22} which contains the MQM annotated data for three translation directions: En-De, Zh-En, and En-Ru.
Reference A (refA) is used as the standard reference.
The quality of references can affect the performance of reference-based metrics (see Appendix~\ref{sec:ref_quality}).
The golden quality score is calculated from the MQM ratings annotated by humans.
The weighting scheme of each error severity and category can be found in \citet{DBLP:journals/tacl/FreitagFGRTM21}.

\paragraph{Models.} We evaluate both the closed model GPT-3.5-turbo and open models, including the Llama2-Chat series~\cite{Touvron2023Llama2O} and Mistral-7B-Instruct~\cite{Jiang2023Mistral7}.
We only consider the chat version of these models because base models without alignment may not follow instructions according to our preliminary study.
All of these models possess a certain level of translation ability in the specified language pairs~\cite{DBLP:journals/corr/ZhuLDXKCLH23}.

\paragraph{Evaluation Metrics.} Following \citet{DBLP:conf/eamt/KocmiF23}, we use the system-level accuracy and segment-level Kendall's $\tau$ correlation as our primary evaluation metrics, complemented by the Pearson correlation $\rho$.
We use the PERM-BOTH hypothesis test with 1000 resampling runs and p=0.05~\cite{deutsch-etal-2021-statistical}.

\begin{figure*}[h]
    \centering
    \includegraphics[width=.95\linewidth]{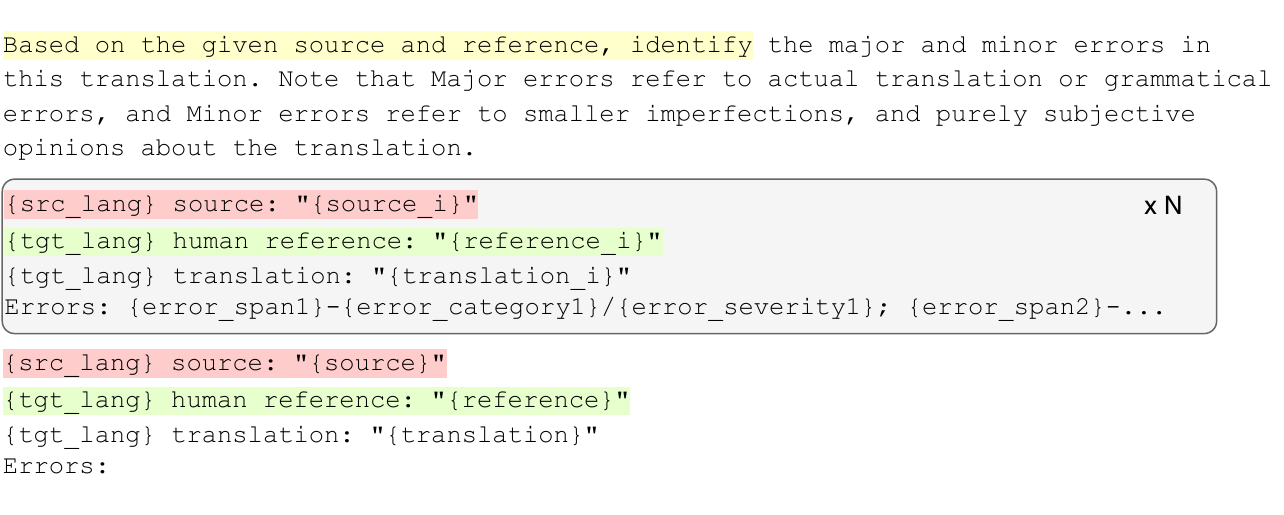}
    \vspace{-10pt}
    \caption{The AutoMQM prompt template. The green parts and red parts are included according to whether the related information is given. The yellow part is also determined by the input mode. The text in the shaded area is an in-context demonstration, followed by the test sample. Detailed prompts can be found in Appendix~\ref{sec:detailofautomqm}.}
    \label{fig:automqm}
\end{figure*}

\subsection{Results}
\label{sec:coarse_results}
Table~\ref{tab:coarse} demonstrates the main results of the meta-evaluation of the coarse-grained translation quality score, in which we include COMET-22~\cite{DBLP:conf/wmt/ReiSAZFGLCM22}, BLEU and chrF as baselines.

\begin{table}[t]
    \setlength{\tabcolsep}{2pt}
    \TABLEFONT
    \centering
    \begin{tabular}{lccccc}
        \toprule
        Model & Part & Acc. & En-De $\tau$ & Zh-En $\tau$ & En-Ru $\tau$ \\
        \midrule
        \multirow{2}{*}{GPT-3.5-turbo} & src & 0.051 & 0.001 & -0.010 & -0.009 \\
        & ref & 0.066 & 0.073 & 0.056 & 0.025 \\
        \hdashline
        \multirow{2}{*}{Llama2-7B-Chat} & src & -0.030 & -0.046 & -0.028 & -0.012 \\
        & ref & 0.159 & 0.181 & 0.163 & 0.193 \\
        \hdashline
        \multirow{2}{*}{Llama2-13B-Chat} & src & -0.067 & 0.021 & -0.014 & 0.026 \\
        & ref & 0.043 & 0.087 & 0.034 & 0.050 \\
        \hdashline
        \multirow{2}{*}{Llama2-70B-Chat} & src & 0.013 & -0.048 & -0.021 & -0.065 \\
        & ref & 0.093 & 0.067 & 0.056 & 0.023 \\
        \hdashline
        \multirow{2}{*}{Mistral-7B-Instruct} & src & -0.053 & -0.005 & 0.008 & -0.018 \\
        & ref & 0.097 & 0.055 & -0.002 & -0.040 \\
        \bottomrule
    \end{tabular}
    \vspace{-5pt}
    \caption{The Shapley values that quantify the impact of the source and reference parts on the system-level accuracy and Kendall's $\tau$ correlations in the score prediction task across different language pairs.}
    \label{tab:shapley_coarse}
\end{table}

One of the most surprising findings is that the R-T mode is the most effective among the four modes in most cases.
The R-T mode surpasses other modes in system-level accuracy and segment-level correlations across models, particularly excelling with strong models like GPT-3.5 and Llama2-70B-Chat.
This suggests that the reference information can significantly enhance the evaluation accuracy, but the source information has little or no impact on the translation evaluation task.
On the other hand, the numerous lower scores in S-T mode compared to T mode also indicate that LLMs do not fully utilize their cross-lingual capability and may even be confused by the source in this task.
We further calculate the Shapley values~\cite{shapley1953value} that assess the contributions of the source and reference part, as shown in Table~\ref{tab:shapley_coarse}.
Higher number means more positive influence, and vice versa.
A positive number means that the information has a positive effect while a negative number means a negative effect.
The way to calculate Shapley Values can be found in Appendix~\ref{sec:shapley}.
The reference parts contribute more than the source parts, which can even have negative impacts.

There is another unexpected observation that the T mode can achieve much better performance than a random guess.
We posit that LLMs evaluate translations solely based on fluency, which is positively correlated to the translation quality.
Our further analysis by log-probability in Appendix~\ref{log_probability} supports this hypothesis.

Compared to baseline metrics, our findings align with those of \citet{DBLP:conf/eamt/KocmiF23}.
LLMs are better at system-level evaluation but are inferior at segment-level correlations than COMET-22.
Metrics based on strong LLMs outperform both BLEU and chrF.
COMET-22 is built upon the pre-trained encoder language model XLM-R~\cite{DBLP:conf/acl/ConneauKGCWGGOZ20} and it is an ensemble between several models with different input.
It's sequence tagger performs better with the S-R-T mode at the segment level.
We suspect that the underlying mechanism of encoder-only models may differ from decoder-only models.
Moreover, COMET is fine-tuned with a mount of task-specific data including both the source and the reference.
This may also enhance the performance of COMET on this task.

\section{Fine-grained Error Detection}
\label{sec:fine_grained}
While coarse-grained scoring methods have demonstrated their potential, recent innovations like AutoMQM and GEMBA-MQM are eliciting the capabilities of LLMs through the use of specialized prompts, leading to more refined and interpretable results.
We further dive into studying how well LLMs leverage the different information with fine-grained methods.

We adopt the AutoMQM prompt template~\cite{DBLP:journals/corr/FernandesDFRMNGCFF23}, as illustrated in Figure~\ref{fig:automqm}.
The content in the yellow part varies depending on the input mode.
Our assessment encompasses three perspectives: MQM scores, error spans, and error categories, hence offering a comprehensive diagnosis of the model's predictions.

\begin{table}[t]
    \TABLEFONT
    \setlength{\tabcolsep}{2pt}
    \centering
    \begin{tabular}{lcccccc}
    \toprule
         \multirow{2}{*}{Model} & \multirow{2}{*}{Mode}  & 2 LPs & \multicolumn{2}{c}{En-De} & \multicolumn{2}{c}{Zh-En} \\
         \cmidrule{3-7}
         & & Acc. & $\tau$ & $\rho$ & $\tau$ & $\rho$ \\
         \midrule
         \multicolumn{1}{@{}l}{{\footnotesize \textbf{AutoMQM}}} \\
         \cdashlinelr{1-7}
         \multirow{4}{*}{GPT-3.5-turbo} & T & 0.757 & 0.221 & 0.283 & 0.264 & 0.353 \\
         & S-T & 0.751 & 0.150 & 0.222 & 0.289 & 0.394 \\
         & R-T & \textbf{0.858} & 0.275 & 0.331 & \textbf{0.359} & \textbf{0.479} \\
         & S-R-T & 0.769 & \textbf{0.284} & \textbf{0.349} & 0.353 & 0.460 \\
         \cdashlinelr{1-7}
         \multirow{4}{*}{Llama2-7B} & T & 0.556 & \textbf{0.077} & \textbf{0.111} & \textbf{0.106} & \textbf{0.216} \\
         & S-T & \textbf{0.592} & 0.071 & 0.073 & 0.074 & 0.119 \\
         & R-T & 0.527 & \textbf{0.077} & 0.102 & \textbf{0.106} & 0.146 \\
         & S-R-T & 0.533 & 0.063 & 0.075 & 0.086 & 0.133 \\
         \cdashlinelr{1-7}
         \multirow{4}{*}{Llama2-13B} & T & 0.544 & 0.078 & \textbf{0.110} & \textbf{0.130} & \textbf{0.220} \\
         & S-T & 0.515 & 0.063 & 0.060 & 0.108 & 0.214 \\
         & R-T & 0.533 & \textbf{0.083} & 0.086 & 0.108 & 0.178 \\
         & S-R-T & \textbf{0.562} & 0.049 & 0.036 & 0.110 & 0.212 \\
         \cdashlinelr{1-7}
         \multirow{4}{*}{Llama2-70B} & T & 0.586 & 0.134 & 0.182 & 0.128 & 0.202 \\
         & S-T & 0.633 & 0.135 & 0.206 & 0.169 & 0.236 \\
         & R-T & 0.627 & \textbf{0.200} & \textbf{0.270} & 0.225 & 0.266 \\
         & S-R-T & \textbf{0.669} & \textbf{0.200} & 0.237 & \textbf{0.248} & \textbf{0.315} \\
         \cdashlinelr{1-7}
         \multirow{4}{*}{Mistral-7B} & T & 0.444 & 0.109 & 0.136 & 0.118 & 0.203 \\
         & S-T & 0.538 & 0.088 & 0.102 & 0.107 & 0.176 \\
         & R-T & \textbf{0.604} & \textbf{0.143} & \textbf{0.185} & 0.116 & 0.190 \\
         & S-R-T & 0.586 & 0.108 & 0.112 & \textbf{0.121} & \textbf{0.212} \\
         \midrule
         \multicolumn{1}{@{}l}{{\footnotesize \textbf{GEMBA}}} \\
         \cdashlinelr{1-7}
         \multirow{4}{*}{GPT-3.5-turbo} & T & 0.728 & 0.264 & 0.272 & 0.229 & 0.223 \\
         & S-T & \textbf{0.852} & 0.247 & 0.226 & 0.188 & 0.211  \\
         & R-T & \textbf{0.852} & 0.273 & 0.290 & \textbf{0.281} & \textbf{0.231} \\
         & S-R-T & 0.828 & \textbf{0.284} & \textbf{0.299} & 0.239 & 0.209 \\
         \cdashlinelr{1-7}
         \multirow{4}{*}{Llama2-70B-Chat} & T & 0.698 & 0.150 & 0.114 & 0.226 & \textbf{0.269} \\
         & S-T & 0.775 & 0.161 & 0.117 & 0.219 & 0.221 \\
         & R-T & \textbf{0.828} & \textbf{0.262} & \textbf{0.222} & \textbf{0.271} & 0.220 \\
         & S-R-T & 0.769 & 0.198 & 0.194 & 0.241 & 0.196 \\
         \midrule
         COMET-22 & / & 0.852 & 0.398 & 0.515 & 0.447 & 0.594 \\
         BLEU & / & 0.556 & 0.167 & 0.212 & 0.077 & 0.123 \\
         chrF & / & 0.592 & 0.217 & 0.267 & 0.098 & 0.099 \\
         \bottomrule
    \end{tabular}
    \vspace{-5pt}
    \caption{The system-level accuracy and segment-level Kendall's $\tau$ and Pearson $\rho$ correlations of AutoMQM with different models. All of the models use the AutoMQM prompt except the last five. The highest scores of different input modes of each model are in bold.}
    \label{tab:fine-grained}
\end{table}

\begin{table*}[t]
    \TABLEFONT
    \centering
    \begin{tabular}{lcccc}
        \toprule
        Model & Mode & SP / SR / SF1 & MP / MR / MF1 & MCC \\
        \midrule
        \multirow{4}{*}{GPT-3.5-turbo} & T & 0.162 / 0.375 / 0.227 & 0.122 / 0.155 / 0.136 & 0.153 \\
        & S-T & 0.237 / 0.207 / 0.221 & 0.192 / 0.192 / 0.192 & 0.150 \\
        & R-T & 0.239 / 0.378 / \textbf{0.293} & 0.202 / 0.344 / \textbf{0.254} & \textbf{0.208} \\
        & S-R-T & 0.214 / 0.354 / 0.267 & 0.179 / 0.348 / 0.236 & 0.180 \\
        \cdashlinelr{1-5}
        \multirow{4}{*}{Llama2-7B} & T & 0.110 / 0.520 / \textbf{0.181} & 0.056 / 0.414 / \textbf{0.098} & 0.057 \\
        & S-T & 0.085 / 0.329 / 0.135 & 0.041 / 0.243 / 0.070 & \textbf{0.061} \\
        & R-T & 0.112 / 0.309 / 0.165 & 0.056 / 0.219 / 0.090 & 0.045 \\
        & S-R-T & 0.092 / 0.260 / 0.136 & 0.048 / 0.201 / 0.077 & 0.056 \\
        \cdashlinelr{1-5}
        \multirow{4}{*}{Llama2-13B} & T & 0.113 / 0.604 / \textbf{0.191} & 0.055 / 0.503 / 0.100 & \textbf{0.079} \\
        & S-T & 0.084 / 0.448 / 0.141 & 0.037 / 0.351 / 0.067 & 0.051 \\
        & R-T & 0.119 / 0.433 / 0.186 & 0.064 / 0.391 / \textbf{0.110} & 0.071 \\
        & S-R-T & 0.098 / 0.405 / 0.158 & 0.049 / 0.360 / 0.086 & 0.053 \\
        \cdashlinelr{1-5}
        \multirow{4}{*}{Llama2-70B} & T & 0.107 / 0.665 / 0.185 & 0.056 / 0.646 / 0.106 & 0.065 \\
        & S-T & 0.104 / 0.592 / 0.177 & 0.058 / 0.541 / 0.101 & 0.072 \\
        & R-T & 0.124 / 0.631 / \textbf{0.207} & 0.071 / 0.576 / 0.127 & 0.109 \\
        & S-R-T & 0.121 / 0.659 / 0.204 & 0.072 / 0.577 / \textbf{0.128} & \textbf{0.111} \\
        \cdashlinelr{1-5}
        \multirow{4}{*}{Mistral-7B} & T & 0.108 / 0.679 / \textbf{0.186} & 0.054 / 0.639 / 0.099 & \textbf{0.069} \\
        & S-T & 0.101 / 0.569 / 0.171 & 0.051 / 0.546 / 0.094 & 0.056 \\
        & R-T & 0.108 / 0.537 / 0.179 & 0.056 / 0.524 / \textbf{0.101} & 0.058 \\
        & S-R-T & 0.105 / 0.545 / 0.177 & 0.052 / 0.520 / 0.094 & 0.055 \\
        \bottomrule
    \end{tabular}
    \vspace{-5pt}
    \caption{The results of span meta-evaluation. All of the scores are micro-averaged across two language directions. The highest F1 scores and MCC are in bold.}
    \label{tab:span}
\end{table*}

\subsection{Experimental Setup}
\paragraph{Data.} We sample a portion of the WMT22 test set as our test set due to limited budgets (see Appendix~\ref{sec:token}).
Specifically, we uniformly sample 200 source sentences and all corresponding system outputs from the test set.
There are 16 systems with MQM scores in the En-De and Zh-En directions, resulting in a total of 3200 samples for each direction.
Following~\citet{DBLP:journals/corr/FernandesDFRMNGCFF23}, the in-context demonstrations are sampled from the data in WMT21 Metric Shared Task~\cite{DBLP:conf/wmt/FreitagRMLSFLB21}.
The number of in-context demonstrations is 4 and stratified sampling with a set of rejection criteria is used.\footnote{These models have terrible performance using this prompt without demonstrations, as shown in \citet{DBLP:journals/corr/FernandesDFRMNGCFF23}}
Since there are no MQM ratings for the En-Ru direction in the WMT21 dataset, we only assess the other two directions.

\paragraph{Models.}
We evaluate the GPT-3.5-turbo and the Llama2 base series.
In our preliminary study, the Llama2 chat models cannot follow the output format in this prompt.
Therefore, we decide to assess the base models only.
All models in this experiment generate text using greedy decoding.

\paragraph{Meta Evaluation.}
Based on the identified error categories and severity, we compute an MQM score for each sample according to Google's MQM error weighting~\cite{DBLP:journals/tacl/FreitagFGRTM21}.
Since we do not predict sub-categories, we only assign a score of $-5$ for a major error and $-1$ for a minor error.
We adopt the previous metrics to evaluate the MQM scores.

We also assess the quality of the identified error spans.
Similar to \citet{DBLP:journals/corr/FernandesDFRMNGCFF23}, we calculate the precision, recall, F1 score, and Matthews Correlation Coefficient (MCC) for the predicted error spans.
In particular, given the gold error spans $S=\{e_1, \dots, e_n\}$, $e_j=\{w_i, w_{i+1},\dots\}$ denotes each error span containing the wrong words, where $w_i$ is the $i$-th word in the sentence.
The span of each error is $P(e_j)=\{i|w_i\in e_j\}$.
Then we count the span overlap based on the set $P(S)=\bigcup_{j=1}^n P(e_j)$.
The span precision (SP) and span recall (SR) of the predicted error spans $\hat{S}$ are defined as follows:
\begin{equation}
    \mathrm{SP}=\frac{\lvert P(S)\cap P(\hat{S})\rvert}{\lvert P(\hat{S})\rvert}
\end{equation}
\begin{equation}
    \mathrm{SR}=\frac{\lvert P(S)\cap P(\hat{S})\rvert}{\lvert P(S)\rvert}
\end{equation}
The span F1 score (SF1) is the harmonic mean of SP and SR.
Since major errors contribute most to the quality score, we calculate the major precision (MP) and major recall (MR) as follows:
\begin{equation}
    \mathrm{MP}=\frac{\lvert P(S_{maj})\cap P(\hat{S}_{maj})\rvert}{\lvert P(\hat{S}_{maj})\rvert}
\end{equation}
\begin{equation}
    \mathrm{MR}=\frac{\lvert P(S_{maj})\cap P(\hat{S}_{maj})\rvert}{\lvert P(S_{maj})\rvert}
\end{equation}
where $S_{maj}\subseteq S$ is the subset only containing major errors, and major F1 (MF1) score is the harmonic mean.
Note that our MR is slightly different from \citeposs{DBLP:journals/corr/FernandesDFRMNGCFF23} MR, which takes into account both minor and major prediction errors.
In this way, we can better evaluate the performance of predicting the major errors.

In addition, we calculate the precision, recall, and F1 score for the error category.
Specifically, let $\text{Cat}(e)$ denote the error category, and $\text{Cat}(S)=(\text{Cat}(e_1),\dots,\text{Cat}(e_n))$ denote the gold labels. The function $\text{Count}(S, c)$ calculates the count of occurrences of category $c$ within $\text{Cat}(S)$.
The precision and recall of the category $c$ are defined as:
\begin{equation}
    \mathrm{P_c}=\frac{\min(\text{Count}(S,c), \text{Count}(\hat{S},c))}{\text{Count}(\hat{S},c)}
\end{equation}
\begin{equation}
    \mathrm{R_c}=\frac{\min(\text{Count}(S,c), \text{Count}(\hat{S},c))}{\text{Count}(S,c)}
\end{equation}
And the $\text{F}1_c$ score is the harmonic mean of the precision and recall.
Here we ignore the sub-categories since AutoMQM does not predict sub-categories.
Note that these three scores only consider the error categories and do not necessitate the correct identification of error positions for simplicity.

\subsection{Results}
\label{sec:fine_results}
\paragraph{Score Meta-evaluation.}
Table~\ref{tab:fine-grained} shows that GPT-3.5, Llama2-70B, and Mistral-7B achieve the best or second-best score with the R-T mode, suggesting a limitation in their ability to employ cross-lingual capabilities for this task.
However, while the T mode appears to yield the strong results for weak models such as Llama2-13B and Llama2-7B, it is important to note that their overall performance remains substantially low.
This leads to the hypothesis that these weak models may not fully understand the task, thereby failing to effectively identify errors.
Nevertheless, the contribution of the reference is much larger than that of the source in most cases, as shown in Table~\ref{tab:shapley_fine}.
These results also indicate the limitation of cross-lingual capabilities of LLMs to evaluate translations.

\begin{table}[t]
    \TABLEFONT
    \centering
    \begin{tabular}{lcccc}
        \toprule
        Model & Part & Acc. & En-De $\tau$ & Zh-En $\tau$ \\
        \midrule
        \multirow{2}{*}{GPT-3.5-turbo} & src & -0.047 & -0.031 & 0.009 \\
        & ref & 0.059 & 0.094 & 0.079 \\
        \hdashline
        \multirow{2}{*}{Llama2-7B} & src & 0.021 & -0.010 & -0.026 \\
        & ref & -0.044 & -0.004 & 0.006 \\
        \hdashline
        \multirow{2}{*}{Llama2-13B} & src & 0.000 & -0.025 & -0.010 \\
        & ref & 0.018 & -0.004 & -0.010 \\
        \hdashline
        \multirow{2}{*}{Llama2-70B} & src & 0.045 & 0.001 & 0.032 \\
        & ref & 0.039 & 0.066  & 0.088 \\
        \hdashline
        \multirow{2}{*}{Mistral-7B} & src & 0.038 & -0.028 & -0.003 \\
        & ref & 0.104 & 0.027 & 0.006 \\
        \bottomrule
    \end{tabular}
    \vspace{-5pt}
    \caption{The Shapley values in the error detection task across different language pairs.}
    \label{tab:shapley_fine}
\end{table}

\begin{table*}[t]
    \TABLEFONT
    \centering
    \setlength{\tabcolsep}{3pt}
    \begin{tabular}{lccccccc}
        \toprule
        Model & Mode & Accuracy & Fluency & Terminology & Style & Locale & No-Error \\
        \midrule
        \multirow{4}{*}{GPT-3.5-turbo} & T & 0.31/0.25/0.28 & 0.21/0.14/\textbf{0.17} & 0.03/0.05/0.04 & 0.13/0.29/\textbf{0.18} & 0.00/0.00/0.00 & 0.57/0.80/0.67 \\
        & S-T & 0.45/0.19/0.26 & 0.27/0.06/0.10 & 0.04/0.02/0.02 & 0.15/0.05/0.07 & 0.00/0.00/0.00 & 0.54/0.94/0.69 \\
        & R-T & 0.43/0.41/\textbf{0.42} & 0.27/0.09/0.13 & 0.03/0.04/0.03 & 0.17/0.17/0.17 & 0.00/0.00/0.00 & 0.60/0.84/\textbf{0.70} \\
        & S-R-T & 0.43/0.41/\textbf{0.42} & 0.25/0.08/0.13 & 0.04/0.05/\textbf{0.05} & 0.18/0.17/\textbf{0.18} & 0.00/0.00/0.00 & 0.61/0.81/0.69 \\
        \cdashlinelr{1-8}
        \multirow{4}{*}{Llama2-7B} & T & 0.20/0.66/0.31 & 0.16/0.09/\textbf{0.12} & 0.01/0.11/\textbf{0.03} & 0.12/0.12/\textbf{0.12} & 0.05/0.04/\textbf{0.05} & 0.57/0.04/0.08 \\
        & S-T & 0.21/0.65/\textbf{0.32} & 0.16/0.08/0.10 & 0.01/0.11/0.02 & 0.12/0.12/\textbf{0.12} & 0.02/0.04/0.03 & 0.54/0.06/0.10 \\
        & R-T & 0.24/0.50/\textbf{0.32} & 0.19/0.08/\textbf{0.12} & 0.01/0.08/0.02 & 0.12/0.10/0.11 & 0.03/0.02/0.02 & 0.56/0.22/\textbf{0.32} \\
        & S-R-T & 0.22/0.52/0.31 & 0.16/0.06/0.08 & 0.02/0.11/\textbf{0.03} & 0.10/0.08/0.09 & 0.02/0.02/0.02 & 0.54/0.21/0.30 \\
        \cdashlinelr{1-8}
        \multirow{4}{*}{Llama2-13B} & T & 0.20/0.34/0.25 & 0.13/0.47/0.20 & 0.01/0.04/0.01 & 0.09/0.20/0.12 & 0.00/0.00/0.00 & 0.61/0.03/0.05 \\
        & S-T & 0.20/0.39/0.27 & 0.12/0.40/0.19 & 0.01/0.06/0.02 & 0.08/0.14/0.11 & 0.00/0.00/0.00 & 0.51/0.03/0.05 \\
        & R-T & 0.25/0.34/\textbf{0.29} & 0.16/0.39/\textbf{0.22} & 0.02/0.06/\textbf{0.03} & 0.11/0.15/\textbf{0.13} & 0.00/0.00/0.00 & 0.55/0.14/\textbf{0.22} \\
        & S-R-T & 0.24/0.37/\textbf{0.29} & 0.15/0.37/0.21 & 0.01/0.05/0.02 & 0.10/0.10/0.10 & 0.00/0.00/0.00 & 0.53/0.13/0.20 \\
        \cdashlinelr{1-8}
        \multirow{4}{*}{Llama2-70B} & T & 0.17/0.49/0.26 & 0.12/0.36/0.18 & 0.01/0.08/0.02 & 0.08/0.08/0.08 & 0.00/0.00/0.00 & 0.65/0.03/0.05 \\
        & S-T & 0.19/0.53/0.28 & 0.12/0.34/0.18 & 0.01/0.08/0.02 & 0.09/0.11/0.10 & 0.02/0.04/\textbf{0.03} & 0.70/0.08/0.15 \\
        & R-T & 0.22/0.56/\textbf{0.32} & 0.13/0.30/0.18 & 0.02/0.06/\textbf{0.03} & 0.11/0.13/\textbf{0.12} & 0.05/0.02/\textbf{0.03} & 0.70/0.14/\textbf{0.23} \\
        & S-R-T & 0.23/0.54/\textbf{0.32} & 0.12/0.36/0.18 & 0.01/0.04/0.01 & 0.10/0.13/0.11  & 0.02/0.02/0.02 & 0.74/0.13/0.22 \\
        \cdashlinelr{1-8}
        \multirow{4}{*}{Mistral-7B} & T & 0.16/0.48/0.24 & 0.11/0.38/0.17 & 0.02/0.03/\textbf{0.02} & 0.08/0.13/\textbf{0.10} & 0.05/0.04/\textbf{0.04} & 0.63/0.03/0.06 \\
        & S-T & 0.17/0.56/0.26 & 0.12/0.31/0.17 & 0.00/0.00/0.00 & 0.07/0.12/0.09 & 0.04/0.02/0.03 & 0.54/0.04/0.07 \\
        & R-T & 0.20/0.52/\textbf{0.29} & 0.14/0.34/\textbf{0.20} & 0.00/0.00/0.00 & 0.08/0.10/0.09 & 0.04/0.02/0.03 & 0.58/0.10/\textbf{0.16} \\
        & S-R-T & 0.20/0.53/\textbf{0.29} & 0.14/0.32/\textbf{0.20} & 0.00/0.00/0.00 & 0.09/0.09/0.09 & 0.05/0.02/0.03 & 0.60/0.09/\textbf{0.16} \\
        \bottomrule
    \end{tabular}
    \vspace{-5pt}
    \caption{The results of the category evaluation. The numbers in each cell are in the format of $P_c/R_c/F1_c$, where $c$ is the category in the column header. Locale stands for the Locale Convention error category. }
    \label{tab:category}
\end{table*}

Besides, we also find something different with the conclusions of \citet{DBLP:journals/corr/FernandesDFRMNGCFF23}.
The AutoMQM prompt outperforms the GEMBA-SQM prompt when using GPT-3.5-Turbo, which is consistent with the previous work using PaLM2~\cite{Anil2023PaLM2T}, but this trend does not extend to the Llama2 series (the 7B and 13B models have a similar phenomenon).
Due to the lack of training details on PaLM2, we speculate that PaLM2 may have a larger model scale and enhanced multilingual capabilities than the Llama2 series.

\paragraph{Span Meta-evaluation.}
The results of the span meta-evaluation presented in Table~\ref{tab:span} demonstrate a similar pattern to the score meta-evaluation. 
GPT-3.5-Turbo and Llama2-70B have better performance when using the R-T mode, while small models are not stable.
Overall, the performance of the R-T mode still surpasses that of both the S-T and S-R-T modes.
This suggests that the limitations in the cross-lingual capabilities of LLMs also exist in word-level translation evaluation tasks.

Unexpectedly, it appears that identifying major errors poses a greater challenge.
The MF1 scores are consistently lower than the corresponding SF1 scores across all tested models.
All of these models exhibit an apparently low level of performance, suggesting substantial room for progress in the error span prediction.

\paragraph{Category Meta-evaluation.}
Table~\ref{tab:category} presents the outcomes of the category meta-evaluation.
When scrutinizing the F1 scores across the models for each designated category, the overall performance is very poor despite the simplification.
Notably, the scores of the Accuracy category surpass those of other categories, with GPT-3.5 demonstrating a relative advantage over the Llama2 models, particularly when provided with a reference.

The AutoMQM prompt lacks explicit definitions for each category, requiring the models to infer the meaning of each from the demonstrations.
The Accuracy category predominates all other categories except the No-Error category.
This prevalence likely biases the model towards a more frequent prediction of Accuracy errors.
The remaining categories exhibit diminished F1 scores, which are attributed to the models' limited understanding of the inherent semantics associated with each category due to their low frequency.
This pattern persists irrespective of different models or input modes.

``No-Error'' is a special category, as it is mutually exclusive with other error categories.
For analytical simplicity, it is treated analogously to a category, with F1 scores computed accordingly.
In this regard, GPT-3.5 exhibits a pronounced competence in identifying error-free samples in stark contrast to the Llama2 models.
Weak models exhibit a propensity for overestimating the presence of errors.

\begin{table}[t]
    \TABLEFONT
    \centering
    \begin{tabular}{lcccc}
        \toprule
        Model & SP & SR & SF1 & accuracy \\
        \midrule
        GPT-3.5-turbo & 0.186 & 0.626 & 0.287 & 0.553 \\
        Llama2-7B & 0.109 & 0.296 & 0.160 & 0.293 \\
        Llama2-13B & 0.137 & 0.500 & 0.215 & 0.513 \\
        Llama2-70B & 0.147 & 0.836 & 0.250 & 0.753 \\
        \bottomrule
    \end{tabular}
    \vspace{-5pt}
    \caption{The results of critical error detection. Each experiment is run with three different random seeds.}
    \label{tab:critical_error}
\end{table}

\subsection{Critical Error Detection}
To better understand the cross-lingual ability of LLMs, we investigate whether they can detect the critical translation errors that are easy to discover.
We extract 50 samples from the test set of WMT22's Critical Error Detection Task~\cite{zerva-etal-2022-findings}.
Specifically, we only use the ``BAD'' samples from the En-De subset and manually label one critical error span for each sample.
The samples with omission errors are excluded, keeping the addition errors, named entity errors, negation errors and number errors.
We use the AutoMQM prompt with the S-T mode to determine whether LLMs can utilize the source information to identify the critical error spans.
SP, SR, SF1 and accuracy are used to measure the performance.
The accuracy here is calculated as the ratio of how many critical error spans are completely identified.

The results are demonstrated in Table~\ref{tab:critical_error}.
Strong models like GPT-3.5 and Llama2-70B can identify most errors.
However, the precision is very low, indicating that they tend to over-predict errors.
On the other hand, there remains a noticeable probability, exceeding 25\%, that they may overlook crucial information in the source.
This suggests that LLMs cannot fully utilize the source information, leading to the failure of error detection.
There some cases shown in Figure~\ref{fig:ced_cases}.

\begin{table}[ht]
    \centering
    \TABLEFONT
    \begin{tabular}{cccccc}
    \toprule
        Direction & \#S-T & \#R-T & \#S-R-T & Total & No-error\%  \\
    \midrule
        En-De & 2940 & 2993 & 3026 & 8959 & 58.7\% \\
        Zh-En & 3342 & 3204 & 3204 & 9750 & 29.3\% \\
        En-De$^\dag$ & 1712 & 1736 & 1812 & 5260 & 29.7\% \\
    \bottomrule
    \end{tabular}
    \caption{The statistics of the training set. \#S-T/\#R-T/\#S-R-T is the number of samples in this mode after random assignment. No-error\% is the No-error rate of samples. En-De$^\dag$ is the down-sampled subset.}
    \label{tab:training_set}
\end{table}

\begin{table*}[t]
    \TABLEFONT
    \centering
    \begin{tabular}{lcccccccc}
    \toprule
         \multirow{2}{*}{Model} & \multirow{2}{*}{Mode}  & All LPs & \multicolumn{2}{c}{En-De} & \multicolumn{2}{c}{Zh-En} & \multicolumn{2}{c}{En-Ru} \\
         \cmidrule{3-9}
         & & Acc. & $\tau$ & $\rho$ & $\tau$ & $\rho$ & $\tau$ & $\rho$ \\
         \midrule
         \multirow{3}{*}{Fine-tuned Llama2-7B} & S-T & 0.832 & 0.072 & 0.080 & 0.368 & 0.453 & 0.181 & 0.221 \\
         & R-T & \textbf{0.847} & \textbf{0.139*} & \textbf{0.199*} & \textbf{0.407} & \textbf{0.492} & \textbf{0.259*} & \textbf{0.319*} \\
         & S-R-T & \textbf{0.847} & 0.114 & 0.145 & 0.401 & 0.488 & 0.228 & 0.289 \\
         \midrule
         \multirow{3}{*}{Fine-tuned Llama2-7B$^\dag$} & S-T & \textbf{0.828} & 0.153 & 0.161 & 0.366 & 0.455 & 0.208 & 0.236 \\
         & R-T & 0.818 & \textbf{0.229*} & \textbf{0.218*} & \textbf{0.412*} & \textbf{0.506} & \textbf{0.278*} & \textbf{0.300} \\
         & S-R-T & \textbf{0.828} & 0.199 & 0.201 & 0.403 & 0.503 & 0.242 & 0.285 \\
         \midrule
         GEMBA-Llama2-7B-Chat & R-T & 0.788 & 0.217 & 0.200 & 0.284 & 0.260 & 0.213 & 0.177 \\
         GEMBA-GPT-3.5-turbo & R-T & 0.891 & 0.284 & 0.280 & 0.286 & 0.230 & 0.253 & 0.217 \\
         \bottomrule
    \end{tabular}
    \vspace{-5pt}
    \caption{The system-level accuracy and segment-level Kendall's $\tau$ and Pearson $\rho$ correlations of the fine-tuned Llama2. Fine-tuned Llama2-7B$^\dag$ uses the down-sampled training set. Starred values are significantly better than those of the other two input modes.}
    \label{tab:fine-tuned}
\end{table*}

\section{Fine-tuning LLMs with MQM data}
\label{sec:fine-tune}
We further investigate the effect of fine-tuning an open LLM with task-specific data to determine if it can eliminate the above limitation.
\subsection{Experimental Setup}
In this experiment, we integrate the En-De and Zh-En samples from the WMT21 dataset to form the supervised training set, and employ the WMT22 dataset as the test set.
The organization of the training samples adheres to the Alpaca~\cite{alpaca} instruction template, where the instruction and the input parts are identical with the AutoMQM prompt.
Regarding the output part, our format mirrors that of InstructScore~\cite{Xu2023INSTRUCTSCORETE}, with the exception of the explanation component.\footnote{In our preliminary experiments, we used the output format of AutoMQM, but the results were terrible.}
Note that we also ignore the error sub-category here.
To accommodate the three input modes, i.e. the S-T, R-T and S-R-T mode, each training sample is randomly assigned with one mode.
The statistics of the training set in can be found in Table~\ref{tab:training_set}.
We fine-tune the Llama2-7B base model for 3 epochs, using a decayed learning rate of 2e-5 and a batch size of 128.

\subsection{Results}
We have some interesting findings in Table~\ref{tab:fine-tuned}.
Firstly, the performance of the R-T mode remains significantly superior to that of the other two modes, indicating that the model still cannot make full use of the source information after naive fine-tuning. 
Secondly, The overall performance of the fine-tuned Llama2 is stronger than GEMBA-Llama2-7B-Chat, proving the effectiveness of further fine-tuning for this task.
However, the distribution of the training data is crucial.
The fine-tuned model outperforms GPT-3.5-turbo on Zh-En segment-level correlations with the proper data distribution.
On the contrary, the performance on En-De direction degrades due to the extremely imbalanced En-De training data, where samples with No-error dominates, as shown in Table~\ref{tab:training_set}.
To mitigate the problem of unbalanced distribution, we down-sample the No-error samples in the En-De corpus and keep about 30\% No-error samples.
The Zh-En samples remain unchanged.
Fine-tuning with more balanced data can effectively enhance the En-De segment-level correlations, as shown in Table~\ref{tab:fine-tuned}.
The change of En-De data also brings benefits to other directions.
Lastly, it is noteworthy that the evaluation capability, to some extent, can be transferred to the language pair not encountered in the fine-tuning stage.
The fine-tuned model achieves even higher correlations without seeing any En-Ru samples, compared to GEMBA-Llama2-7B-Chat.

\section{Conclusion}
We empirically analyze how well LLMs incorporate the source and reference information for translation evaluation, comparing the effectiveness of open and closed LLMs through prompting and fine-tuning.
Our results reveal their limitations in fully exploiting the cross-lingual capability for the task, with the inclusion of source information even occasionally proving detrimental to performance.
Furthermore, our work contributes a detailed meta-evaluation of spans and categories with the fine-grained evaluation method, along with the critical error detection task.
These findings not only furnish insights into the current capabilities and limitations of LLMs in translation evaluation, but also establish a foundational basis for subsequent scholarly endeavors.
In the future, we would like to extend these analyses to other NLG evaluation tasks.

\section{Limitations}
We discuss the limitations and future research directions of our work in this section.
\begin{itemize}
    \item In experiments, we mainly use the prompts from the previous works~\cite{DBLP:journals/corr/FernandesDFRMNGCFF23,DBLP:conf/eamt/KocmiF23}.
    These prompts are may not the best prompt that can fully elicit the ability of LLMs on this task.
    It's important to note that our conclusion may not apply to all prompts.
    However, the current popular prompts that simply ask LLMs to predict scores or fine-grained errors can be negatively affected by the source.
    Designing prompts that can better elicit the cross-lingual capability of LLMs is a topic for future research.
    \item We do not evaluate other closed LLMs like GPT-4 due to the limited resources. The tokens consumed in our experiments are recorded in the Appendix~\ref{sec:token}.
    We leave assessing additional LLMs with more test data as future work.
    \item We do not dive into how to better fine-tune the open model. More carefully designed training data or pipelines may bring greater improvement for this task.
    \item In this work, we only focus on the translation evaluation task which is a sub-field of NLG evaluation tasks.
    Future research should focus on extending these analyses to other NLG evaluation tasks.
\end{itemize}

\section*{Acknowledgements}
We would like to thank the anonymous reviewers for their insightful comments. Shujian Huang and Zhirui Zhang are the corresponding authors. This work is supported by National Science Foundation of China (No. 62376116, 62176120), the Liaoning Provincial Research Foundation for Basic Research (No. 2022-KF-26-02), research project of Nanjing University-China Mobile Joint Institute.

\bibliography{anthology,custom}

\begin{thebibliography}{45}
\expandafter\ifx\csname natexlab\endcsname\relax\def\natexlab#1{#1}\fi

\bibitem[{Anil et~al.(2023)Anil, Dai, Firat, Johnson, Lepikhin, Passos, Shakeri, Taropa, Bailey, Chen, Chu, Clark, Shafey, Huang, Meier-Hellstern, Mishra, Moreira, Omernick, Robinson, Ruder, Tay, Xiao, Xu, Zhang, Abrego, Ahn, Austin, Barham, Botha, Bradbury, Brahma, Brooks, Catasta, Cheng, Cherry, Choquette-Choo, Chowdhery, Cr{\'e}py, Dave, Dehghani, Dev, Devlin, D'iaz, Du, Dyer, Feinberg, Feng, Fienber, Freitag, Garc{\'i}a, Gehrmann, Gonz{\'a}lez, Gur-Ari, Hand, Hashemi, Hou, Howland, Hu, Hui, Hurwitz, Isard, Ittycheriah, Jagielski, Jia, Kenealy, Krikun, Kudugunta, Lan, Lee, Lee, Li, Li, Li, Li, Li, Lim, Lin, Liu, Liu, Maggioni, Mahendru, Maynez, Misra, Moussalem, Nado, Nham, Ni, Nystrom, Parrish, Pellat, Polacek, Polozov, Pope, Qiao, Reif, Richter, Riley, Ros, Roy, Saeta, Samuel, Shelby, Slone, Smilkov, So, Sohn, Tokumine, Valter, Vasudevan, Vodrahalli, Wang, Wang, Wang, Wang, Wieting, Wu, Xu, Xu, Xue, Yin, Yu, Zhang, Zheng, Zheng, Zhou, Zhou, Petrov, and Wu}]{Anil2023PaLM2T}
Rohan Anil, Andrew~M. Dai, Orhan Firat, Melvin Johnson, Dmitry Lepikhin, Alexandre~Tachard Passos, Siamak Shakeri, Emanuel Taropa, Paige Bailey, Z.~Chen, Eric Chu, J.~Clark, Laurent~El Shafey, Yanping Huang, Kathleen~S. Meier-Hellstern, Gaurav Mishra, Erica Moreira, Mark Omernick, Kevin Robinson, Sebastian Ruder, Yi~Tay, Kefan Xiao, Yuanzhong Xu, Yujing Zhang, Gustavo~Hernandez Abrego, Junwhan Ahn, Jacob Austin, Paul Barham, Jan~A. Botha, James Bradbury, Siddhartha Brahma, Kevin~Michael Brooks, Michele Catasta, Yongzhou Cheng, Colin Cherry, Christopher~A. Choquette-Choo, Aakanksha Chowdhery, C~Cr{\'e}py, Shachi Dave, Mostafa Dehghani, Sunipa Dev, Jacob Devlin, M.~C. D'iaz, Nan Du, Ethan Dyer, Vladimir Feinberg, Fan Feng, Vlad Fienber, Markus Freitag, Xavier Garc{\'i}a, Sebastian Gehrmann, Lucas Gonz{\'a}lez, Guy Gur-Ari, Steven Hand, Hadi Hashemi, Le~Hou, Joshua Howland, An~Ren Hu, Jeffrey Hui, Jeremy Hurwitz, Michael Isard, Abe Ittycheriah, Matthew Jagielski, Wen~Hao Jia, Kathleen Kenealy, Maxim Krikun,
  Sneha Kudugunta, Chang Lan, Katherine Lee, Benjamin Lee, Eric Li, Mu-Li Li, Wei Li, Yaguang Li, Jun~Yu Li, Hyeontaek Lim, Han Lin, Zhong-Zhong Liu, Frederick Liu, Marcello Maggioni, Aroma Mahendru, Joshua Maynez, Vedant Misra, Maysam Moussalem, Zachary Nado, John Nham, Eric Ni, Andrew Nystrom, Alicia Parrish, Marie Pellat, Martin Polacek, Alex Polozov, Reiner Pope, Siyuan Qiao, Emily Reif, Bryan Richter, Parker Riley, Alexandra Ros, Aurko Roy, Brennan Saeta, Rajkumar Samuel, Renee~Marie Shelby, Ambrose Slone, Daniel Smilkov, David~R. So, Daniela Sohn, Simon Tokumine, Dasha Valter, Vijay Vasudevan, Kiran Vodrahalli, Xuezhi Wang, Pidong Wang, Zirui Wang, Tao Wang, John Wieting, Yuhuai Wu, Ke~Xu, Yunhan Xu, Lin~Wu Xue, Pengcheng Yin, Jiahui Yu, Qiaoling Zhang, Steven Zheng, Ce~Zheng, Wei Zhou, Denny Zhou, Slav Petrov, and Yonghui Wu. 2023.
\newblock \href {https://doi.org/10.48550/arXiv.2305.10403} {Palm 2 technical report}.
\newblock \emph{ArXiv}, abs/2305.10403.

\bibitem[{Bahdanau et~al.(2015)Bahdanau, Cho, and Bengio}]{DBLP:journals/corr/BahdanauCB14}
Dzmitry Bahdanau, Kyunghyun Cho, and Yoshua Bengio. 2015.
\newblock \href {http://arxiv.org/abs/1409.0473} {Neural machine translation by jointly learning to align and translate}.
\newblock In \emph{3rd International Conference on Learning Representations, {ICLR} 2015, San Diego, CA, USA, May 7-9, 2015, Conference Track Proceedings}.

\bibitem[{Banerjee and Lavie(2005)}]{DBLP:conf/acl/BanerjeeL05}
Satanjeev Banerjee and Alon Lavie. 2005.
\newblock \href {https://aclanthology.org/W05-0909/} {{METEOR:} an automatic metric for {MT} evaluation with improved correlation with human judgments}.
\newblock In \emph{Proceedings of the Workshop on Intrinsic and Extrinsic Evaluation Measures for Machine Translation and/or Summarization@ACL 2005, Ann Arbor, Michigan, USA, June 29, 2005}, pages 65--72. Association for Computational Linguistics.

\bibitem[{Chan et~al.(2023)Chan, Chen, Su, Yu, Xue, Zhang, Fu, and Liu}]{Chan2023ChatEvalTB}
Chi-Min Chan, Weize Chen, Yusheng Su, Jianxuan Yu, Wei Xue, Shan Zhang, Jie Fu, and Zhiyuan Liu. 2023.
\newblock \href {https://doi.org/10.48550/arXiv.2308.07201} {Chateval: Towards better llm-based evaluators through multi-agent debate}.
\newblock \emph{ArXiv}, abs/2308.07201.

\bibitem[{Conneau et~al.(2020)Conneau, Khandelwal, Goyal, Chaudhary, Wenzek, Guzm{\'{a}}n, Grave, Ott, Zettlemoyer, and Stoyanov}]{DBLP:conf/acl/ConneauKGCWGGOZ20}
Alexis Conneau, Kartikay Khandelwal, Naman Goyal, Vishrav Chaudhary, Guillaume Wenzek, Francisco Guzm{\'{a}}n, Edouard Grave, Myle Ott, Luke Zettlemoyer, and Veselin Stoyanov. 2020.
\newblock \href {https://doi.org/10.18653/V1/2020.ACL-MAIN.747} {Unsupervised cross-lingual representation learning at scale}.
\newblock In \emph{Proceedings of the 58th Annual Meeting of the Association for Computational Linguistics, {ACL} 2020, Online, July 5-10, 2020}, pages 8440--8451. Association for Computational Linguistics.

\bibitem[{Deutsch et~al.(2021)Deutsch, Dror, and Roth}]{deutsch-etal-2021-statistical}
Daniel Deutsch, Rotem Dror, and Dan Roth. 2021.
\newblock \href {https://doi.org/10.1162/tacl_a_00417} {A statistical analysis of summarization evaluation metrics using resampling methods}.
\newblock \emph{Transactions of the Association for Computational Linguistics}, 9:1132--1146.

\bibitem[{Devlin et~al.(2019)Devlin, Chang, Lee, and Toutanova}]{DBLP:conf/naacl/DevlinCLT19}
Jacob Devlin, Ming{-}Wei Chang, Kenton Lee, and Kristina Toutanova. 2019.
\newblock \href {https://doi.org/10.18653/V1/N19-1423} {{BERT:} pre-training of deep bidirectional transformers for language understanding}.
\newblock In \emph{Proceedings of the 2019 Conference of the North American Chapter of the Association for Computational Linguistics: Human Language Technologies, {NAACL-HLT} 2019, Minneapolis, MN, USA, June 2-7, 2019, Volume 1 (Long and Short Papers)}, pages 4171--4186. Association for Computational Linguistics.

\bibitem[{Fernandes et~al.(2023)Fernandes, Deutsch, Finkelstein, Riley, Martins, Neubig, Garg, Clark, Freitag, and Firat}]{DBLP:journals/corr/FernandesDFRMNGCFF23}
Patrick Fernandes, Daniel Deutsch, Mara Finkelstein, Parker Riley, Andr{\'{e}} F.~T. Martins, Graham Neubig, Ankush Garg, Jonathan~H. Clark, Markus Freitag, and Orhan Firat. 2023.
\newblock \href {https://doi.org/10.48550/ARXIV.2308.07286} {The devil is in the errors: Leveraging large language models for fine-grained machine translation evaluation}.
\newblock \emph{ArXiv}, abs/2308.07286.

\bibitem[{Freitag et~al.(2021{\natexlab{a}})Freitag, Foster, Grangier, Ratnakar, Tan, and Macherey}]{DBLP:journals/tacl/FreitagFGRTM21}
Markus Freitag, George~F. Foster, David Grangier, Viresh Ratnakar, Qijun Tan, and Wolfgang Macherey. 2021{\natexlab{a}}.
\newblock \href {https://doi.org/10.1162/TACL\_A\_00437} {Experts, errors, and context: {A} large-scale study of human evaluation for machine translation}.
\newblock \emph{Trans. Assoc. Comput. Linguistics}, 9:1460--1474.

\bibitem[{Freitag et~al.(2023)Freitag, Mathur, Lo, Avramidis, Rei, Thompson, Kocmi, Blain, Deutsch, Stewart, Zerva, Castilho, Lavie, and Foster}]{freitag-etal-2023-results}
Markus Freitag, Nitika Mathur, Chi-kiu Lo, Eleftherios Avramidis, Ricardo Rei, Brian Thompson, Tom Kocmi, Frederic Blain, Daniel Deutsch, Craig Stewart, Chrysoula Zerva, Sheila Castilho, Alon Lavie, and George Foster. 2023.
\newblock \href {https://aclanthology.org/2023.wmt-1.51} {Results of {WMT}23 metrics shared task: Metrics might be guilty but references are not innocent}.
\newblock In \emph{Proceedings of the Eighth Conference on Machine Translation}, pages 578--628, Singapore. Association for Computational Linguistics.

\bibitem[{Freitag et~al.(2022)Freitag, Rei, Mathur, Lo, Stewart, Avramidis, Kocmi, Foster, Lavie, and Martins}]{DBLP:conf/wmt/FreitagRMLSAKFLM22}
Markus Freitag, Ricardo Rei, Nitika Mathur, Chi{-}kiu Lo, Craig Stewart, Eleftherios Avramidis, Tom Kocmi, George~F. Foster, Alon Lavie, and Andr{\'{e}} F.~T. Martins. 2022.
\newblock \href {https://aclanthology.org/2022.wmt-1.2} {Results of {WMT22} metrics shared task: Stop using {BLEU} - neural metrics are better and more robust}.
\newblock In \emph{Proceedings of the Seventh Conference on Machine Translation, {WMT} 2022, Abu Dhabi, United Arab Emirates (Hybrid), December 7-8, 2022}, pages 46--68. Association for Computational Linguistics.

\bibitem[{Freitag et~al.(2021{\natexlab{b}})Freitag, Rei, Mathur, Lo, Stewart, Foster, Lavie, and Bojar}]{DBLP:conf/wmt/FreitagRMLSFLB21}
Markus Freitag, Ricardo Rei, Nitika Mathur, Chi{-}kiu Lo, Craig Stewart, George~F. Foster, Alon Lavie, and Ondrej Bojar. 2021{\natexlab{b}}.
\newblock \href {https://aclanthology.org/2021.wmt-1.73} {Results of the {WMT21} metrics shared task: Evaluating metrics with expert-based human evaluations on {TED} and news domain}.
\newblock In \emph{Proceedings of the Sixth Conference on Machine Translation, WMT@EMNLP 2021, Online Event, November 10-11, 2021}, pages 733--774. Association for Computational Linguistics.

\bibitem[{Fu et~al.(2023)Fu, Ng, Jiang, and Liu}]{Fu2023GPTScoreEA}
Jinlan Fu, See-Kiong Ng, Zhengbao Jiang, and Pengfei Liu. 2023.
\newblock \href {https://doi.org/10.48550/arXiv.2302.04166} {Gptscore: Evaluate as you desire}.
\newblock \emph{ArXiv}, abs/2302.04166.

\bibitem[{Guerreiro et~al.(2023)Guerreiro, Rei, van Stigt, Coheur, Colombo, and Martins}]{Guerreiro2023xCOMETTM}
Nuno~M. Guerreiro, Ricardo Rei, Daan van Stigt, Lu{\'i}sa Coheur, Pierre Colombo, and Andr{\'e} Martins. 2023.
\newblock \href {https://doi.org/10.48550/arXiv.2310.10482} {xcomet: Transparent machine translation evaluation through fine-grained error detection}.
\newblock \emph{ArXiv}, abs/2310.10482.

\bibitem[{Hassan et~al.(2018)Hassan, Aue, Chen, Chowdhary, Clark, Federmann, Huang, Junczys-Dowmunt, Lewis, Li et~al.}]{hassan2018achieving}
Hany Hassan, Anthony Aue, Chang Chen, Vishal Chowdhary, Jonathan Clark, Christian Federmann, Xuedong Huang, Marcin Junczys-Dowmunt, William Lewis, Mu~Li, et~al. 2018.
\newblock \href {http://arxiv.org/abs/1803.05567} {Achieving human parity on automatic chinese to english news translation}.
\newblock \emph{arXiv preprint arXiv:1803.05567}.

\bibitem[{Jiang et~al.(2023)Jiang, Sablayrolles, Mensch, Bamford, Chaplot, de~Las~Casas, Bressand, Lengyel, Lample, Saulnier, Lavaud, Lachaux, Stock, Scao, Lavril, Wang, Lacroix, and Sayed}]{Jiang2023Mistral7}
Albert~Qiaochu Jiang, Alexandre Sablayrolles, Arthur Mensch, Chris Bamford, Devendra~Singh Chaplot, Diego de~Las~Casas, Florian Bressand, Gianna Lengyel, Guillaume Lample, Lucile Saulnier, L'elio~Renard Lavaud, Marie-Anne Lachaux, Pierre Stock, Teven~Le Scao, Thibaut Lavril, Thomas Wang, Timoth{\'e}e Lacroix, and William~El Sayed. 2023.
\newblock \href {https://doi.org/10.48550/arXiv.2310.06825} {Mistral 7b}.
\newblock \emph{ArXiv}, abs/2310.06825.

\bibitem[{Kocmi and Federmann(2023{\natexlab{a}})}]{DBLP:journals/corr/KocmiF23}
Tom Kocmi and Christian Federmann. 2023{\natexlab{a}}.
\newblock \href {https://doi.org/10.48550/arXiv.2310.13988} {{GEMBA-MQM:} detecting translation quality error spans with {GPT-4}}.
\newblock \emph{ArXiv}, abs/2310.13988.

\bibitem[{Kocmi and Federmann(2023{\natexlab{b}})}]{DBLP:conf/eamt/KocmiF23}
Tom Kocmi and Christian Federmann. 2023{\natexlab{b}}.
\newblock \href {https://aclanthology.org/2023.eamt-1.19} {Large language models are state-of-the-art evaluators of translation quality}.
\newblock In \emph{Proceedings of the 24th Annual Conference of the European Association for Machine Translation, {EAMT} 2023, Tampere, Finland, 12-15 June 2023}, pages 193--203. European Association for Machine Translation.

\bibitem[{Liu et~al.(2023{\natexlab{a}})Liu, Iter, Xu, Wang, Xu, and Zhu}]{Liu2023GEvalNE}
Yang Liu, Dan Iter, Yichong Xu, Shuo Wang, Ruochen Xu, and Chenguang Zhu. 2023{\natexlab{a}}.
\newblock \href {https://doi.org/10.48550/arXiv.2303.16634} {G-eval: Nlg evaluation using gpt-4 with better human alignment}.
\newblock \emph{ArXiv}, abs/2303.16634.

\bibitem[{Liu et~al.(2019)Liu, Ott, Goyal, Du, Joshi, Chen, Levy, Lewis, Zettlemoyer, and Stoyanov}]{DBLP:journals/corr/LiuOGDJ19}
Yinhan Liu, Myle Ott, Naman Goyal, Jingfei Du, Mandar Joshi, Danqi Chen, Omer Levy, Mike Lewis, Luke Zettlemoyer, and Veselin Stoyanov. 2019.
\newblock \href {http://arxiv.org/abs/1907.11692} {Roberta: {A} robustly optimized {BERT} pretraining approach}.
\newblock \emph{ArXiv}, abs/1907.11692.

\bibitem[{Liu et~al.(2023{\natexlab{b}})Liu, Yang, Huang, Zhang, Huang, Wei, Deng, Sun, and Zhang}]{Liu2023CalibratingLE}
Yuxuan Liu, Tianchi Yang, Shaohan Huang, Zihan Zhang, Haizhen Huang, Furu Wei, Weiwei Deng, Feng Sun, and Qi~Zhang. 2023{\natexlab{b}}.
\newblock \href {https://doi.org/10.48550/arXiv.2309.13308} {Calibrating llm-based evaluator}.
\newblock \emph{ArXiv}, abs/2309.13308.

\bibitem[{Lu et~al.(2023)Lu, Qiu, Ding, Xie, and Tao}]{DBLP:journals/corr/LuQDXT23}
Qingyu Lu, Baopu Qiu, Liang Ding, Liping Xie, and Dacheng Tao. 2023.
\newblock \href {https://doi.org/10.48550/ARXIV.2303.13809} {Error analysis prompting enables human-like translation evaluation in large language models: {A} case study on chatgpt}.
\newblock \emph{ArXiv}, abs/2303.13809.

\bibitem[{OpenAI(2023)}]{DBLP:journals/corr/gpt4}
OpenAI. 2023.
\newblock \href {https://doi.org/10.48550/ARXIV.2303.08774} {{GPT-4} technical report}.
\newblock \emph{ArXiv}, abs/2303.08774.

\bibitem[{Papineni et~al.(2002)Papineni, Roukos, Ward, and Zhu}]{DBLP:conf/acl/PapineniRWZ02}
Kishore Papineni, Salim Roukos, Todd Ward, and Wei{-}Jing Zhu. 2002.
\newblock \href {https://aclanthology.org/P02-1040/} {Bleu: a method for automatic evaluation of machine translation}.
\newblock In \emph{Proceedings of the 40th Annual Meeting of the Association for Computational Linguistics, July 6-12, 2002, Philadelphia, PA, {USA}}, pages 311--318. {ACL}.

\bibitem[{Popovic(2015)}]{DBLP:conf/wmt/Popovic15}
Maja Popovic. 2015.
\newblock \href {https://doi.org/10.18653/v1/w15-3049} {chrf: character n-gram f-score for automatic {MT} evaluation}.
\newblock In \emph{Proceedings of the Tenth Workshop on Statistical Machine Translation, WMT@EMNLP 2015, 17-18 September 2015, Lisbon, Portugal}, pages 392--395. The Association for Computer Linguistics.

\bibitem[{Rei et~al.(2022)Rei, de~Souza, Alves, Zerva, Farinha, Glushkova, Lavie, Coheur, and Martins}]{DBLP:conf/wmt/ReiSAZFGLCM22}
Ricardo Rei, Jos{\'{e}} G.~C. de~Souza, Duarte~M. Alves, Chrysoula Zerva, Ana~C. Farinha, Taisiya Glushkova, Alon Lavie, Lu{\'{\i}}sa Coheur, and Andr{\'{e}} F.~T. Martins. 2022.
\newblock \href {https://aclanthology.org/2022.wmt-1.52} {{COMET-22:} unbabel-ist 2022 submission for the metrics shared task}.
\newblock In \emph{Proceedings of the Seventh Conference on Machine Translation, {WMT} 2022, Abu Dhabi, United Arab Emirates (Hybrid), December 7-8, 2022}, pages 578--585. Association for Computational Linguistics.

\bibitem[{Rei et~al.(2020)Rei, Stewart, Farinha, and Lavie}]{DBLP:conf/emnlp/ReiSFL20}
Ricardo Rei, Craig Stewart, Ana~C. Farinha, and Alon Lavie. 2020.
\newblock \href {https://doi.org/10.18653/V1/2020.EMNLP-MAIN.213} {{COMET:} {A} neural framework for {MT} evaluation}.
\newblock In \emph{Proceedings of the 2020 Conference on Empirical Methods in Natural Language Processing, {EMNLP} 2020, Online, November 16-20, 2020}, pages 2685--2702. Association for Computational Linguistics.

\bibitem[{Sellam et~al.(2020)Sellam, Das, and Parikh}]{DBLP:conf/acl/SellamDP20}
Thibault Sellam, Dipanjan Das, and Ankur~P. Parikh. 2020.
\newblock \href {https://doi.org/10.18653/V1/2020.ACL-MAIN.704} {{BLEURT:} learning robust metrics for text generation}.
\newblock In \emph{Proceedings of the 58th Annual Meeting of the Association for Computational Linguistics, {ACL} 2020, Online, July 5-10, 2020}, pages 7881--7892. Association for Computational Linguistics.

\bibitem[{Shapley(1953)}]{shapley1953value}
Lloyd~S. Shapley. 1953.
\newblock A value for n-person games.
\newblock \emph{Contributions to the Theory of Games}, pages 307--317.

\bibitem[{Taori et~al.(2023)Taori, Gulrajani, Zhang, Dubois, Li, Guestrin, Liang, and Hashimoto}]{alpaca}
Rohan Taori, Ishaan Gulrajani, Tianyi Zhang, Yann Dubois, Xuechen Li, Carlos Guestrin, Percy Liang, and Tatsunori~B. Hashimoto. 2023.
\newblock Stanford alpaca: An instruction-following llama model.
\newblock \url{https://github.com/tatsu-lab/stanford_alpaca}.

\bibitem[{Thompson and Post(2020)}]{DBLP:conf/emnlp/ThompsonP20}
Brian Thompson and Matt Post. 2020.
\newblock \href {https://doi.org/10.18653/v1/2020.emnlp-main.8} {Automatic machine translation evaluation in many languages via zero-shot paraphrasing}.
\newblock In \emph{Proceedings of the 2020 Conference on Empirical Methods in Natural Language Processing, {EMNLP} 2020, Online, November 16-20, 2020}, pages 90--121. Association for Computational Linguistics.

\bibitem[{Touvron et~al.(2023)Touvron, Martin, Stone, Albert, Almahairi, Babaei, Bashlykov, Batra, Bhargava, Bhosale, Bikel, Blecher, Ferrer, Chen, Cucurull, Esiobu, Fernandes, Fu, Fu, Fuller, Gao, Goswami, Goyal, Hartshorn, Hosseini, Hou, Inan, Kardas, Kerkez, Khabsa, Kloumann, Korenev, Koura, Lachaux, Lavril, Lee, Liskovich, Lu, Mao, Martinet, Mihaylov, Mishra, Molybog, Nie, Poulton, Reizenstein, Rungta, Saladi, Schelten, Silva, Smith, Subramanian, Tan, Tang, Taylor, Williams, Kuan, Xu, Yan, Zarov, Zhang, Fan, Kambadur, Narang, Rodriguez, Stojnic, Edunov, and Scialom}]{Touvron2023Llama2O}
Hugo Touvron, Louis Martin, Kevin~R. Stone, Peter Albert, Amjad Almahairi, Yasmine Babaei, Nikolay Bashlykov, Soumya Batra, Prajjwal Bhargava, Shruti Bhosale, Daniel~M. Bikel, Lukas Blecher, Cristian~Cant{\'o}n Ferrer, Moya Chen, Guillem Cucurull, David Esiobu, Jude Fernandes, Jeremy Fu, Wenyin Fu, Brian Fuller, Cynthia Gao, Vedanuj Goswami, Naman Goyal, Anthony~S. Hartshorn, Saghar Hosseini, Rui Hou, Hakan Inan, Marcin Kardas, Viktor Kerkez, Madian Khabsa, Isabel~M. Kloumann, A.~V. Korenev, Punit~Singh Koura, Marie-Anne Lachaux, Thibaut Lavril, Jenya Lee, Diana Liskovich, Yinghai Lu, Yuning Mao, Xavier Martinet, Todor Mihaylov, Pushkar Mishra, Igor Molybog, Yixin Nie, Andrew Poulton, Jeremy Reizenstein, Rashi Rungta, Kalyan Saladi, Alan Schelten, Ruan Silva, Eric~Michael Smith, R.~Subramanian, Xia Tan, Binh Tang, Ross Taylor, Adina Williams, Jian~Xiang Kuan, Puxin Xu, Zhengxu Yan, Iliyan Zarov, Yuchen Zhang, Angela Fan, Melanie Kambadur, Sharan Narang, Aurelien Rodriguez, Robert Stojnic, Sergey Edunov, and
  Thomas Scialom. 2023.
\newblock \href {https://api.semanticscholar.org/CorpusID:259950998} {Llama 2: Open foundation and fine-tuned chat models}.
\newblock \emph{ArXiv}, abs/2307.09288.

\bibitem[{Vaswani et~al.(2017)Vaswani, Shazeer, Parmar, Uszkoreit, Jones, Gomez, Kaiser, and Polosukhin}]{DBLP:conf/nips/VaswaniSPUJGKP17}
Ashish Vaswani, Noam Shazeer, Niki Parmar, Jakob Uszkoreit, Llion Jones, Aidan~N. Gomez, Lukasz Kaiser, and Illia Polosukhin. 2017.
\newblock \href {https://proceedings.neurips.cc/paper/2017/hash/3f5ee243547dee91fbd053c1c4a845aa-Abstract.html} {Attention is all you need}.
\newblock In \emph{Advances in Neural Information Processing Systems 30: Annual Conference on Neural Information Processing Systems 2017, December 4-9, 2017, Long Beach, CA, {USA}}, pages 5998--6008.

\bibitem[{Wan et~al.(2022)Wan, Liu, Yang, Zhang, Chen, Wong, and Chao}]{DBLP:conf/acl/0004LYZCWC22}
Yu~Wan, Dayiheng Liu, Baosong Yang, Haibo Zhang, Boxing Chen, Derek~F. Wong, and Lidia~S. Chao. 2022.
\newblock \href {https://doi.org/10.18653/V1/2022.ACL-LONG.558} {Unite: Unified translation evaluation}.
\newblock In \emph{Proceedings of the 60th Annual Meeting of the Association for Computational Linguistics (Volume 1: Long Papers), {ACL} 2022, Dublin, Ireland, May 22-27, 2022}, pages 8117--8127. Association for Computational Linguistics.

\bibitem[{Wang et~al.(2023{\natexlab{a}})Wang, Liang, Meng, Shi, Li, Xu, Qu, and Zhou}]{Wang2023IsCA}
Jiaan Wang, Yunlong Liang, Fandong Meng, Haoxiang Shi, Zhixu Li, Jinan Xu, Jianfeng Qu, and Jie Zhou. 2023{\natexlab{a}}.
\newblock \href {https://arxiv.org/pdf/2303.04048.pdf} {Is chatgpt a good nlg evaluator? a preliminary study}.
\newblock \emph{ArXiv}, abs/2303.04048.

\bibitem[{Wang et~al.(2023{\natexlab{b}})Wang, Lyu, Ji, Zhang, Yu, Shi, and Tu}]{DBLP:conf/emnlp/WangLJZY0T23}
Longyue Wang, Chenyang Lyu, Tianbo Ji, Zhirui Zhang, Dian Yu, Shuming Shi, and Zhaopeng Tu. 2023{\natexlab{b}}.
\newblock \href {https://aclanthology.org/2023.emnlp-main.1036} {Document-level machine translation with large language models}.
\newblock In \emph{Proceedings of the 2023 Conference on Empirical Methods in Natural Language Processing, {EMNLP} 2023, Singapore, December 6-10, 2023}, pages 16646--16661. Association for Computational Linguistics.

\bibitem[{Wang et~al.(2023{\natexlab{c}})Wang, Li, Chen, Zhu, Lin, Cao, Liu, Liu, and Sui}]{Wang2023LargeLM}
Peiyi Wang, Lei Li, Liang Chen, Dawei Zhu, Binghuai Lin, Yunbo Cao, Qi~Liu, Tianyu Liu, and Zhifang Sui. 2023{\natexlab{c}}.
\newblock \href {https://doi.org/10.48550/arXiv.2305.17926} {Large language models are not fair evaluators}.
\newblock \emph{ArXiv}, abs/2305.17926.

\bibitem[{West et~al.(2023)West, Lu, Dziri, Brahman, Li, Hwang, Jiang, Fisher, Ravichander, Chandu, Newman, Koh, Ettinger, and Choi}]{DBLP:journals/corr/West2023TheGA}
Peter West, Ximing Lu, Nouha Dziri, Faeze Brahman, Linjie Li, Jena~D. Hwang, Liwei Jiang, Jillian Fisher, Abhilasha Ravichander, Khyathi Chandu, Benjamin Newman, Pang~Wei Koh, Allyson Ettinger, and Yejin Choi. 2023.
\newblock \href {https://doi.org/10.48550/arXiv.2311.00059} {The generative {AI} paradox: "what it can create, it may not understand"}.
\newblock \emph{ArXiv}, abs/2311.00059.

\bibitem[{Xu et~al.(2023)Xu, Wang, Pan, Song, Freitag, Wang, and Li}]{Xu2023INSTRUCTSCORETE}
Wenda Xu, Danqing Wang, Liangming Pan, Zhenqiao Song, Markus Freitag, William~Yang Wang, and Lei Li. 2023.
\newblock \href {https://doi.org/10.48550/ARXIV.2305.14282} {{INSTRUCTSCORE:} towards explainable text generation evaluation with automatic feedback}.
\newblock \emph{ArXiv}, abs/2305.14282.

\bibitem[{Yan et~al.(2023)Yan, Wang, Zhao, Huang, Chen, and Wang}]{DBLP:conf/acl/YanWZHCW23}
Yiming Yan, Tao Wang, Chengqi Zhao, Shujian Huang, Jiajun Chen, and Mingxuan Wang. 2023.
\newblock \href {https://doi.org/10.18653/V1/2023.ACL-LONG.297} {{BLEURT} has universal translations: An analysis of automatic metrics by minimum risk training}.
\newblock In \emph{Proceedings of the 61st Annual Meeting of the Association for Computational Linguistics (Volume 1: Long Papers), {ACL} 2023, Toronto, Canada, July 9-14, 2023}, pages 5428--5443. Association for Computational Linguistics.

\bibitem[{Yuan et~al.(2021)Yuan, Neubig, and Liu}]{DBLP:conf/nips/YuanNL21}
Weizhe Yuan, Graham Neubig, and Pengfei Liu. 2021.
\newblock \href {https://proceedings.neurips.cc/paper/2021/hash/e4d2b6e6fdeca3e60e0f1a62fee3d9dd-Abstract.html} {Bartscore: Evaluating generated text as text generation}.
\newblock In \emph{Advances in Neural Information Processing Systems 34: Annual Conference on Neural Information Processing Systems 2021, NeurIPS 2021, December 6-14, 2021, virtual}, pages 27263--27277.

\bibitem[{Zerva et~al.(2022)Zerva, Blain, Rei, Lertvittayakumjorn, C.~de Souza, Eger, Kanojia, Alves, Or{\u{a}}san, Fomicheva, Martins, and Specia}]{zerva-etal-2022-findings}
Chrysoula Zerva, Fr{\'e}d{\'e}ric Blain, Ricardo Rei, Piyawat Lertvittayakumjorn, Jos{\'e}~G. C.~de Souza, Steffen Eger, Diptesh Kanojia, Duarte Alves, Constantin Or{\u{a}}san, Marina Fomicheva, Andr{\'e} F.~T. Martins, and Lucia Specia. 2022.
\newblock \href {https://aclanthology.org/2022.wmt-1.3} {Findings of the {WMT} 2022 shared task on quality estimation}.
\newblock In \emph{Proceedings of the Seventh Conference on Machine Translation (WMT)}, pages 69--99, Abu Dhabi, United Arab Emirates (Hybrid). Association for Computational Linguistics.

\bibitem[{Zhang et~al.(2020)Zhang, Kishore, Wu, Weinberger, and Artzi}]{DBLP:conf/iclr/ZhangKWWA20}
Tianyi Zhang, Varsha Kishore, Felix Wu, Kilian~Q. Weinberger, and Yoav Artzi. 2020.
\newblock \href {https://openreview.net/forum?id=SkeHuCVFDr} {Bertscore: Evaluating text generation with {BERT}}.
\newblock In \emph{8th International Conference on Learning Representations, {ICLR} 2020, Addis Ababa, Ethiopia, April 26-30, 2020}. OpenReview.net.

\bibitem[{Zhao et~al.(2019)Zhao, Peyrard, Liu, Gao, Meyer, and Eger}]{DBLP:conf/emnlp/ZhaoPLGME19}
Wei Zhao, Maxime Peyrard, Fei Liu, Yang Gao, Christian~M. Meyer, and Steffen Eger. 2019.
\newblock \href {https://doi.org/10.18653/V1/D19-1053} {Moverscore: Text generation evaluating with contextualized embeddings and earth mover distance}.
\newblock In \emph{Proceedings of the 2019 Conference on Empirical Methods in Natural Language Processing and the 9th International Joint Conference on Natural Language Processing, {EMNLP-IJCNLP} 2019, Hong Kong, China, November 3-7, 2019}, pages 563--578. Association for Computational Linguistics.

\bibitem[{Zhu et~al.(2023)Zhu, Liu, Dong, Xu, Kong, Chen, Li, and Huang}]{DBLP:journals/corr/ZhuLDXKCLH23}
Wenhao Zhu, Hongyi Liu, Qingxiu Dong, Jingjing Xu, Lingpeng Kong, Jiajun Chen, Lei Li, and Shujian Huang. 2023.
\newblock \href {https://doi.org/10.48550/ARXIV.2304.04675} {Multilingual machine translation with large language models: Empirical results and analysis}.
\newblock \emph{ArXiv}, abs/2304.04675.

\end{thebibliography}

\appendix

\section{GEMBA-SQM Prompts with Four Input Modes}
\label{sec:detailofgemba}
\paragraph{T mode}
\begin{quote}
\small
\texttt{Score the following translation from \{src\_lang\} to \{tgt\_lang\} on a continuous scale from 0 to 100 that starts on ``No meaning preserved'', goes through ``Some meaning preserved'', then ``Most meaning preserved and few grammar mistakes'', up to ``Perfect meaning and grammar''.}

\texttt{\{tgt\_lang\} translation: ``\{translation\}''}\\
\texttt{Score (0-100):}
\end{quote}

\paragraph{S-T mode}
\begin{quote}
\small
\texttt{Score the following translation from \{src\_lang\} to \{tgt\_lang\} on a continuous scale from 0 to 100 that starts on ``No meaning preserved'', goes through ``Some meaning preserved'', then ``Most meaning preserved and few grammar mistakes'', up to ``Perfect meaning and grammar''.}

\texttt{\{src\_lang\} source: ``\{source\}''}\\
\texttt{\{tgt\_lang\} translation: ``\{translation\}''}\\
\texttt{Score (0-100):}
\end{quote}

\paragraph{R-T mode}
\begin{quote}
\small
\texttt{Score the following translation from \{src\_lang\} to \{tgt\_lang\} with respect to the human reference on a continuous scale from 0 to 100 that starts on ``No meaning preserved'', goes through ``Some meaning preserved'', then ``Most meaning preserved and few grammar mistakes'', up to ``Perfect meaning and grammar''.}

\texttt{\{tgt\_lang\} human reference: ``\{reference\}''}\\
\texttt{\{tgt\_lang\} translation: ``\{translation\}''}\\
\texttt{Score (0-100):}
\end{quote}

\paragraph{S-R-T mode}
\begin{quote}
\small
\texttt{Score the following translation from \{src\_lang\} to \{tgt\_lang\} with respect to the human reference on a continuous scale from 0 to 100 that starts on ``No meaning preserved'', goes through ``Some meaning preserved'', then ``Most meaning preserved and few grammar mistakes'', up to ``Perfect meaning and grammar''.}

\texttt{\{src\_lang\} source: ``\{source\}''}\\
\texttt{\{tgt\_lang\} human reference: ``\{reference\}''}\\
\texttt{\{tgt\_lang\} translation: ``\{translation\}''}\\
\texttt{Score (0-100):}
\end{quote}

\section{AutoMQM Prompts with Four Input Modes}
\label{sec:detailofautomqm}
\paragraph{T mode}
\begin{quote}
\small
\texttt{Identify the major and minor errors in this translation. Note that Major errors refer to actual translation or grammatical errors, and Minor errors refer to smaller imperfections, and purely subjective opinions about the translation.}

\texttt{\{tgt\_lang\} translation: ``\{translation\textsubscript{i}\}''}\\
\texttt{Errors: ...}

\texttt{\{tgt\_lang\} translation: ``\{translation\}''}\\
\texttt{Errors:}
\end{quote}

\paragraph{S-T mode}
\begin{quote}
\small
\texttt{Based on the given source, identify the major and minor errors in this translation. Note that Major errors refer to actual translation or grammatical errors, and Minor errors refer to smaller imperfections, and purely subjective opinions about the translation.}

\texttt{\{src\_lang\} source: ``\{source\textsubscript{i}\}''}\\
\texttt{\{tgt\_lang\} translation: ``\{translation\textsubscript{i}\}''}\\
\texttt{Errors: ...}

\texttt{\{src\_lang\} source: ``\{source\}''}\\
\texttt{\{tgt\_lang\} translation: ``\{translation\}''}\\
\texttt{Errors:}
\end{quote}

\paragraph{R-T mode}
\begin{quote}
\small
\texttt{Based on the given reference, identify the major and minor errors in this translation. Note that Major errors refer to actual translation or grammatical errors, and Minor errors refer to smaller imperfections, and purely subjective opinions about the translation.}

\texttt{\{tgt\_lang\} human reference: ``\{reference\textsubscript{i}\}''}\\
\texttt{\{tgt\_lang\} translation: ``\{translation\textsubscript{i}\}''}\\
\texttt{Errors: ...}

\texttt{\{tgt\_lang\} human reference: ``\{reference\}''}\\
\texttt{\{tgt\_lang\} translation: ``\{translation\}''}\\
\texttt{Errors:}
\end{quote}

\paragraph{S-R-T mode}
\begin{quote}
\small
\texttt{Based on the given source and reference, identify the major and minor errors in this translation. Note that Major errors refer to actual translation or grammatical errors, and Minor errors refer to smaller imperfections, and purely subjective opinions about the translation.}

\texttt{\{src\_lang\} source: ``\{source\textsubscript{i}\}''}\\
\texttt{\{tgt\_lang\} human reference: ``\{reference\textsubscript{i}\}''}\\
\texttt{\{tgt\_lang\} translation: ``\{translation\textsubscript{i}\}''}\\
\texttt{Errors: ...}

\texttt{\{src\_lang\} source: ``\{source\}''}\\
\texttt{\{tgt\_lang\} human reference: ``\{reference\}''}\\
\texttt{\{tgt\_lang\} translation: ``\{translation\}''}\\
\texttt{Errors:}
\end{quote}

\section{Effects of Reference Quality}
\label{sec:ref_quality}
According to the results of WMT23 Metrics Shared Task~\cite{freitag-etal-2023-results}, poor human-generated reference translations can dramatically hurt the performance and reliability of the reference-based metrics.
Here we perform a simple experiment to confirm this conclusion.
We extract all of the samples whose MQM score of refA is less than or equal to -2.0 from the WMT22 Zh-En test set, and finally get 5488 samples with 343 different sources
Then we evaluate the performance of GEMBA-SQM-GPT-3.5-turbo and GEMBA-SQM-Llama2-70B-Chat on this test set.
The results are shown in Table~\ref{tab:ref_quality}.
The gap between S-T and R-T/S-R-T gets much smaller.
Sometimes S-T is even better than R-T.
Consequently, we believe that the low-quality references have a negative impact on reference-based methods.

\begin{table}[ht]
    \TABLEFONT
    \centering
    \begin{tabular}{lcccc}
    \toprule
         Model& Mode  & Acc. & $\tau$ & $\rho$ \\
         \midrule
         \multirow{4}{*}{GPT-3.5-turbo}& T & 0.879 & 0.196 & 0.129 \\
         & S-T & 0.890 & 0.142 & 0.149 \\
         & R-T & 0.879 & 0.169 & 0.142 \\
         & S-R-T & 0.789 & 0.187 & 0.183 \\
         \cdashlinelr{1-5}
         \multirow{4}{*}{Llama2-70B-Chat} & T & 0.879 & 0.188 & 0.189 \\
         & S-T & 0.802 & 0.144 & 0.102 \\
         & R-T & 0.824 & 0.179 & 0.147 \\
         & S-R-T & 0.802 & 0.135 & 0.108 \\
         \bottomrule
    \end{tabular}
    \vspace{-5pt}
    \caption{The performance of different models using different input modes on the test set with inaccurate references.}
    \label{tab:ref_quality}
\end{table}

\section{Shapley Values Calculation}
\label{sec:shapley}
We denote the meta-evaluation scores of each input mode as $S_{T}$, $S_{ST}$, $S_{RT}$ and $S_{SRT}$.
The Shapley Value of the source part is
\begin{equation*}
    Shapley_{src}=\frac{(S_{ST}-S_{T})+(S_{SRT}-S_{RT})}{2}.
\end{equation*}
Similarly, the Shapley Value of the reference is
\begin{equation*}
    Shapley_{ref}=\frac{(S_{RT}-S_{T})+(S_{SRT}-S_{ST})}{2}.
\end{equation*}

\section{Analysis by Log-Probability}
\label{log_probability}
We hypothesize that the non-trivial outcomes observed when employing the T mode may be attributed to the LLMs basing their scoring on the quality of the translation sentence provided that the translation is semantically similar to the source.
We measure the quality of a sentence using log-probability.
Moreover, drawing inspiration from generation-based methods, we also calculate the log-probability of the translation as a scoring metric when providing either the source, the reference, or a combination of both.

In this experiment, we only test the open models including both chat and base versions since the log-probability of ChatGPT is inaccessible at that time.
We adopt the same prompt as above (Figure~\ref{fig:gemba-sqm}) for the chat models and just compute the vanilla log-probability of the translation part.
As for the base models, considering they may be confused about the instruction, we only use the equal sign ``='' to concatenate the source, reference, and translation sentences.
For example, the prompt template of the S-R-T mode is ``\{source\} = \{reference\} = \{translation\}'', and that of the T mode is simply the translation sentence ``\{translation\}''.
The log-probability of the translation sentence is computed as follows:
\begin{equation}
    P(\mathbf{t}) = \sum_{i=1}^{N}\log p(t_i|\mathbf{c},\mathbf{t}_{<i})
\end{equation}
where $\mathbf{t}$ is the tokens in the translation sentence of length $N$, and $\mathbf{c}$ is the context before the translation in the prompt, such as the instruction, the source, and reference information.

The test set, models, and metrics are identical to those used in the coarse-grained score prediction experiments, except that we add the base models.

\subsection{Results}
As presented in Table~\ref{tab:log_probability}, we have similar observations to the previous experiment.
The superiority of the R-T mode is more prominent in this experiment, irrespective of the model type and size.
This also corroborates that even powerful large language models cannot utilize the source information effectively in the translation evaluation task.
The performance of the T mode which only computes the translation's log-probability, remains significantly higher than random guess.
The system-level accuracy of the T mode even exceeds the S-T and S-R-T mode by a large margin.
These findings provide strong support for our hypothesis, suggesting that it is plausible for models to offer a relatively accurate score solely based on the quality of the translation sentence.

In this table, we also observe that the performance of each metric does not scale up well with the model size, regardless of further alignment.
The system-level accuracy of models within the same base or chat series is comparable, with the 7B model even slightly outperforming the 70B model.
Meanwhile, some of the segment-level correlations, like the correlations of T and S-T mode, are slightly increasing as the model size up.
However, the slope is very gradual.
We speculate that scaling may bring little benefit to the inherently deficient discriminate capability of auto-regressive language models, which is pertinent to the Generative AI Paradox~\cite{DBLP:journals/corr/West2023TheGA}.

When comparing Table~\ref{tab:coarse} and Table~\ref{tab:log_probability}, a peculiar phenomenon is observed that the segment-level correlations of log-probability are much higher than those of the score prediction method, whereas the system-level accuracy is significantly lower.
We leave the reason behind as future work.

\section{ChatGPT Token Usage}
\label{sec:token}
We record the ChatGPT token usage and cost in Table~\ref{tab:token}.

\begin{table}[t]
    \centering
    \TABLEFONT
    \setlength{\tabcolsep}{2pt}
    \begin{tabular}{cccccc}
    \toprule
    \textbf{Prompt} & \textbf{Input Mode} & \textbf{LP} & \textbf{Samples} & \textbf{Tokens} & \textbf{Cost(\$)}\\
    \midrule
    \multirow{12}{*}{GEMBA} & \multirow{3}{*}{S-T} & En-De & 22725 & 2860k & 5.72 \\
                            &                      & Zh-En & 26340 & 4030k & 8.06 \\
                            &                      & En-Ru & 23326 & 3340k & 6.68 \\
    \cmidrule{2-6}
                             & \multirow{3}{*}{S-R-T} & En-De & 22847 & 3970k & 7.94 \\
                             &                        & Zh-En & 26399 & 5280k & 10.56\\
                             &                        & En-Ru & 24058 & 5010k & 10.02\\
    \cmidrule{2-6}
                             & \multirow{3}{*}{R-T} & En-De & 22738 & 3340k & 6.68 \\
                             &                      & Zh-En & 26676 & 3830k & 7.66 \\
                             &                      & En-Ru & 23841 & 4330k & 8.66 \\
    \cmidrule{2-6}
                             & \multirow{3}{*}{T} & En-De & 22719 & 2240k & 4.48 \\
                             &                    & Zh-En & 27454 & 2660k & 5.32 \\
                             &                    & En-Ru & 23260 & 2700k & 5.40 \\
    \cmidrule{2-6}
                             & \multicolumn{2}{c}{Total} & 292383 & 43590k & 87.18 \\
    \midrule
    \multirow{8}{*}{AutoMQM} & \multirow{3}{*}{S-T} & En-De & 3200 & 2450k & 4.90 \\
                             &                      & Zh-En & 3200 & 2700k & 5.40 \\
    \cmidrule{2-6}
                             & \multirow{3}{*}{S-R-T} & En-De & 3200 & 3470k & 6.94 \\
                             &                        & Zh-En & 3200 & 3520k & 7.04 \\
    \cmidrule{2-6}
                             & \multirow{3}{*}{R-T} & En-De & 3200 & 2810k & 5.62 \\
                             &                      & Zh-En & 3200 & 2360k & 4.72 \\
    \cmidrule{2-6}
                             & \multirow{3}{*}{T} & En-De & 3200 & 1800k & 3.60 \\
                             &                    & Zh-En & 3200 & 1550k & 3.10 \\
    \cmidrule{2-6}
                             & \multicolumn{2}{c}{Total} & 25600 & 20660k & 41.32 \\
    \bottomrule
    \end{tabular}
    \vspace{-5pt}
\caption{ChatGPT token usage in the experiments.}
\label{tab:token}
\end{table}

\begin{figure*}[h]
    \centering
    \includegraphics[width=.95\linewidth]{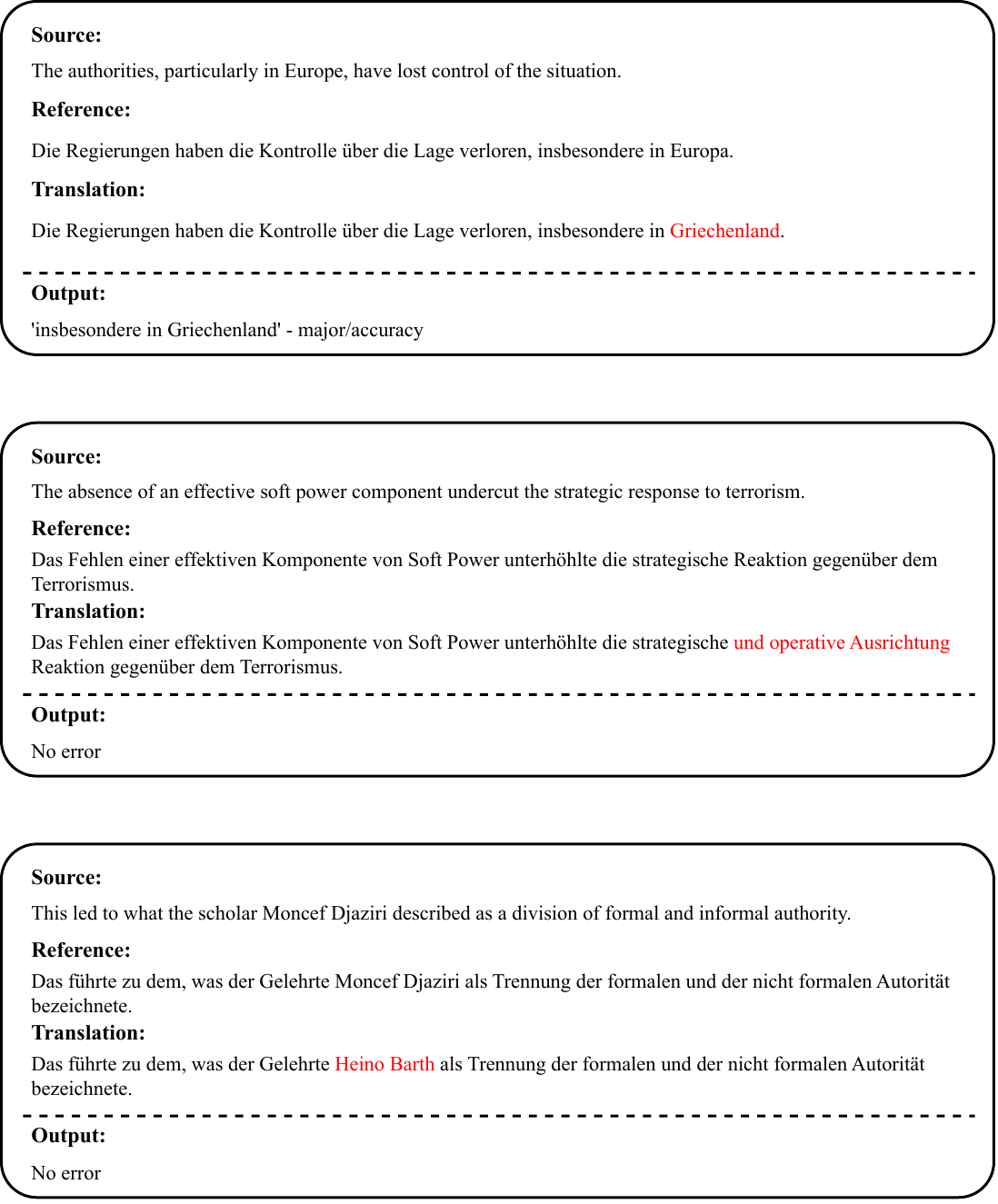}
    \caption{Cases of GPT-3.5's outputs. The texts in red are critical errors. Up: The model identifies the named entity error successfully. Middle: The model fails to detect the addition error. Bottom: The model fails to detect the named entity error. }
    \label{fig:ced_cases}
\end{figure*}

\begin{table*}[t]
    \centering
    \TABLEFONT
    \begin{tabular}{lcccccccc}
    \toprule
         \multirow{2}{*}{Model}& \multirow{2}{*}{Mode}  & All LPs & \multicolumn{2}{c}{En-De} & \multicolumn{2}{c}{Zh-En} & \multicolumn{2}{c}{En-Ru} \\
         \cmidrule{3-9}
         &  & Acc. & $\tau$ & $\rho$ & $\tau$ & $\rho$ & $\tau$ & $\rho$ \\
         \midrule
         \multirow{4}{*}{Llama2-70B-Chat}& T & 0.701 & 0.176 & 0.282 & 0.270 & 0.448 & 0.203 & 0.267\\
         & S-T & 0.485 & 0.168 & 0.297 & 0.290 & 0.466 & 0.187 & 0.252 \\
         & R-T & \textbf{0.730} & \textbf{0.246} & \textbf{0.374} & \textbf{0.333} & \textbf{0.535} & \textbf{0.244} & \textbf{0.331} \\
         & S-R-T & 0.544 & 0.196 & 0.324 & 0.299 & 0.490 & 0.217 & 0.294 \\
         \cdashlinelr{1-9}
         \multirow{4}{*}{Llama2-13B-Chat} & T & 0.693 & 0.172 & 0.276 & 0.269 & 0.444 & 0.199 & 0.262 \\
         & S-T & 0.471 & 0.157 & 0.274 & 0.287 & 0.459 & 0.179 & 0.233 \\
         & R-T & \textbf{0.726} & \textbf{0.238} & \textbf{0.369} & \textbf{0.328} & \textbf{0.531} & \textbf{0.239} & \textbf{0.317} \\
         & S-R-T & 0.620 & 0.200 & 0.331 & 0.293 & 0.486 & 0.215 & 0.283 \\
         \cdashlinelr{1-9}
         \multirow{4}{*}{Llama2-7B-Chat} & T & 0.675 & 0.168 & 0.271 & 0.269 & 0.444 & 0.196 & 0.253 \\
         & S-T & 0.412 & 0.153 & 0.266 & 0.277 & 0.445 & 0.164 & 0.221 \\
         & R-T & \textbf{0.752} & \textbf{0.223} & \textbf{0.350} & \textbf{0.327} & \textbf{0.522} & \textbf{0.231} & \textbf{0.310} \\
         & S-R-T & 0.569 & 0.191 & 0.320 & 0.302 & 0.481 & 0.212 & 0.278 \\
         \cdashlinelr{1-9}
         \multirow{4}{*}{Mistral-7B-Instruct} & T & 0.646 & 0.165 & 0.279 & 0.267 & 0.448 & 0.197 & 0.260 \\
         & S-T & 0.434 & 0.152 & 0.279 & 0.283 & 0.448 & 0.187 & 0.258 \\
         & R-T & \textbf{0.730} & \textbf{0.239} & \textbf{0.374} & \textbf{0.337} & \textbf{0.539} & \textbf{0.243} & \textbf{0.331} \\
         & S-R-T & 0.617 & 0.212 & 0.344 & 0.320 & 0.504 & 0.229 & 0.316 \\
         \cdashlinelr{1-9}
         \multirow{4}{*}{Llama2-70B} & T & 0.708 & 0.185 & 0.295 & 0.284 & 0.458 & 0.219 & 0.282 \\
         & S-T & 0.507 & 0.200 & 0.315 & 0.335 & 0.503 & 0.240 & 0.258 \\
         & R-T & \textbf{0.723} & \textbf{0.256} & \textbf{0.397} & \textbf{0.348} & \textbf{0.548} & \textbf{0.256} & \textbf{0.328} \\
         & S-R-T & 0.591 & 0.221 & 0.352 & 0.348 & 0.524 & 0.244 & 0.279 \\
         \cdashlinelr{1-9}
         \multirow{4}{*}{Llama2-13B} & T & 0.693 & 0.179 & 0.291 & 0.275 & 0.459 & 0.210 & 0.272 \\
         & S-T & 0.460 & 0.188 & 0.297 & 0.327 & 0.496 & 0.224 & 0.242 \\
         & R-T & \textbf{0.726} & \textbf{0.254} & \textbf{0.390} & \textbf{0.349} & \textbf{0.551} & \textbf{0.246} & \textbf{0.319} \\
         & S-R-T & 0.620 & 0.224 & 0.356 & 0.337 & 0.525 & 0.238 & 0.276 \\
         \cdashlinelr{1-9}
         \multirow{4}{*}{Llama2-7B} & T & 0.693 & 0.175 & 0.288 & 0.275 & 0.458 & 0.203 & 0.264 \\
         & S-T & 0.427 & 0.184 & 0.290 & 0.314 & 0.484 & 0.219 & 0.255 \\
         & R-T & \textbf{0.730} & \textbf{0.247} & \textbf{0.377} & \textbf{0.348} & \textbf{0.549} & \textbf{0.244} & \textbf{0.322} \\
         & S-R-T & 0.639 & 0.223 & 0.357 & 0.338 & 0.520 & 0.236 & 0.282 \\
         \cdashlinelr{1-9}
         \multirow{4}{*}{Mistral-7B} & T & 0.682 & 0.179 & 0.285 & 0.278 & 0.462 & 0.211 & 0.267 \\
         & S-T & 0.464 & 0.189 & 0.298 & 0.324 & 0.495 & 0.229 & 0.249 \\
         & R-T & \textbf{0.730} & \textbf{0.252} & \textbf{0.387} & \textbf{0.349} & \textbf{0.551} & \textbf{0.251} & \textbf{0.333} \\
         & S-R-T & 0.664 & 0.223 & 0.359 & 0.341 & 0.533 & 0.245 & 0.296 \\
         \midrule
         COMET-22 & / & 0.839 & 0.368 & 0.512 & 0.428 & 0.585 & 0.400 & 0.469 \\
         BLEU & / & 0.708 & 0.169 & 0.193 & 0.145 & 0.175 & 0.140 & 0.160 \\
         \bottomrule
    \end{tabular}
    \vspace{-5pt}
    \caption{Results of log-probability as a metric on WMT22 test set.}
    \label{tab:log_probability}
\end{table*}

\end{document}